\documentclass[11pt]{article}
\usepackage[preprint]{acl}
\usepackage{times}
\usepackage{latexsym}
\usepackage[T1]{fontenc}
\usepackage[utf8]{inputenc}
\usepackage{microtype}
\usepackage{inconsolata}
\usepackage{booktabs}
\usepackage{graphicx}
\usepackage{float}
\usepackage{multirow}
\usepackage{amsmath}
\usepackage{amssymb}
\usepackage{xcolor}
\usepackage{pgfplots}
\pgfplotsset{compat=1.17}
\usetikzlibrary{positioning,arrows.meta}
\usepackage{listings}
\ifdefined\XeTeXversion
  ="[fonts/DejaVuSansMono.ttf]" at 7pt
\fi
\newcommand{\sys}{TallyMem}

\newcommand{\FullTm}{69.8}

\newcommand{\FullLong}{65.0}

\newcommand{\FullLiteP}{p=0.009}

\newcommand{\FullRuleN}{670}
\newcommand{\FullRuleTm}{64.9}
\newcommand{\FullRuleEv}{68.8}

\newcommand{\FullRuleEvP}{p=0.004}

\newcommand{\FullOtherTm}{71.6}
\newcommand{\FullOtherEv}{71.0}

\newcommand{\FullOtherEvP}{p=0.26}

\newcommand{\FullEvidN}{804}
\newcommand{\FullEvidOtherN}{366}

\newcommand{\FullBinNA}{406}
\newcommand{\FullBinTmA}{77.1}

\newcommand{\FullBinHistA}{77.3}

\newcommand{\FullBinNB}{186}

\newcommand{\FullBinHistB}{68.3}

\newcommand{\FullBinNC}{108}

\newcommand{\FullBinHistC}{61.1}

\newcommand{\FullBinND}{64}

\newcommand{\FullBinHistD}{45.3}

\newcommand{\FullBinNE}{40}
\newcommand{\FullBinTmE}{10.0}

\newcommand{\FullBinHistE}{2.5}

\newcommand{\FullLowTm}{74.3}
\newcommand{\FullLowEv}{71.5}

\newcommand{\FullLowEvP}{p=0.07}
\newcommand{\FullHighN}{212}
\newcommand{\FullHighTm}{43.9}
\newcommand{\FullHighEv}{59.4}

\newcommand{\FullHighEvP}{p<0.001}

\newcommand{\FullBlBm}{46.1}
\newcommand{\FullTmVsBlBmWL}{+763/$-$172}
\newcommand{\FullTmVsBlBmP}{p<0.001}
\newcommand{\FullBlDense}{36.5}

\newcommand{\FullBlRoll}{36.1}

\newcommand{\FullBlMemzero}{24.3}

\newcommand{\FullBlAmem}{38.8}

\newcommand{\FullBlBest}{46.1}

\newcommand{\FullBlAmemScored}{41.2}
\newcommand{\FullBlAmemScoredN}{2,352}
\newcommand{\FullBlMemzeroCalls}{40}
\newcommand{\FullBlAmemCalls}{74}
\newcommand{\FullBlRollCalls}{4}

\newcommand{\FullBlMemzeroFailed}{15}
\newcommand{\FullBlAmemFailedTraj}{12}

\newcommand{\FullBlRollSkipped}{44}
\newcommand{\FullBlRollSkippedTraj}{23}

\newcommand{\FullGate}{71.1}

\newcommand{\FullAgreeN}{688}
\newcommand{\FullAgreePct}{86}
\newcommand{\FullAgreeServer}{64.2}
\newcommand{\FullAgreeDeepinfra}{62.2}
\newcommand{\FullOssLong}{73.6}

\newcommand{\FullOssTm}{78.4}

\newcommand{\FullOssGate}{79.2}

\newcommand{\FullOssLiteP}{p=0.31}

\newcommand{\FullEightLong}{57.6}
\newcommand{\FullEightTmVsLongWL}{+394/$-$315}
\newcommand{\FullEightTmVsLongP}{p=0.003}
\newcommand{\FullEightTm}{60.7}

\newcommand{\FullEightGate}{62.3}
\newcommand{\FullEightGateWL}{+50/$-$12}
\newcommand{\FullEightGateP}{p<0.001}

\newcommand{\MintN}{776}

\newcommand{\MintTm}{35.2}
\newcommand{\MintTally}{49.0}
\newcommand{\MintWL}{+124/$-$17}
\newcommand{\MintP}{p<0.001}

\newcommand{\MintOrdTm}{17.6}
\newcommand{\MintOrdTally}{43.1}

\newcommand{\MintHistTm}{41.0}
\newcommand{\MintHistTally}{56.6}

\newcommand{\MintCountTally}{47.5}

\newcommand{\MintCountLong}{52.5}
\newcommand{\MintLong}{39.2}

\newcommand{\MintTallyVsLongWL}{+164/$-$88}
\newcommand{\MintTallyVsLongP}{p<0.001}

\newcommand{\MintOssLong}{41.1}
\newcommand{\MintOssTm}{37.8}
\newcommand{\MintOssTally}{50.6}

\newcommand{\MintOssTallyVsTmWL}{+124/$-$24}
\newcommand{\MintOssTallyVsTmP}{p<0.001}
\newcommand{\MintOssTallyVsTmCI}{$+$12.9 [$+$9.0, $+$16.8]}
\newcommand{\MintOssTallyVsLongWL}{+166/$-$92}
\newcommand{\MintOssTallyVsLongP}{p<0.001}

\newcommand{\MintOssTmVsLongWL}{+98/$-$124}
\newcommand{\MintOssTmVsLongP}{p=0.09}

\newcommand{\MintEightLong}{36.5}
\newcommand{\MintEightTm}{32.2}
\newcommand{\MintEightTally}{46.1}

\newcommand{\MintEightTallyVsTmWL}{+119/$-$11}
\newcommand{\MintEightTallyVsTmP}{p<0.001}
\newcommand{\MintEightTallyVsTmCI}{$+$13.9 [$+$10.5, $+$17.6]}
\newcommand{\MintEightTallyVsLongWL}{+165/$-$90}
\newcommand{\MintEightTallyVsLongP}{p<0.001}

\newcommand{\MintEightTmVsLongWL}{+94/$-$127}
\newcommand{\MintEightTmVsLongP}{p=0.03}

\newcommand{\MintGainLo}{12.8}
\newcommand{\MintGainHi}{13.9}

\newcommand{\NoCtrTrig}{239}
\newcommand{\NoCtrUntrig}{2,257}
\newcommand{\NoCtrTm}{69.5}
\newcommand{\NoCtrGain}{4.5}
\newcommand{\NoCtrVsLongCI}{$+$4.5 [$+$2.1, $+$6.8]}
\newcommand{\NoCtrVsTmCI}{$-$0.2 [$-$0.8, $+$0.4]}
\newcommand{\NoCtrVsTmWL}{+18/$-$24}
\newcommand{\NoCtrVsTmP}{p=0.44}
\newcommand{\NoCtrSessCI}{$-$1.7 [$-$8.2, $+$4.9]}
\newcommand{\NoCtrSessWL}{+21/$-$25}
\newcommand{\NoCtrSessP}{p=0.66}
\newcommand{\NoCtrAgree}{88.3}
\newcommand{\NoCtrUntrigCI}{$+$3.9 [$+$1.4, $+$6.4]}

\newcommand{\VisibilityLowAcc}{78.9}
\newcommand{\VisibilityHighAcc}{32.2}
\newcommand{\VisibilityLowN}{114}
\newcommand{\VisibilityHighN}{87}
\newcommand{\VisibilityLowCorrect}{90}
\newcommand{\VisibilityHighCorrect}{28}

\newcommand{\CodeN}{630}
\newcommand{\CodeTm}{61.9}
\newcommand{\CodeTally}{66.0}
\newcommand{\CodeAll}{62.1}

\newcommand{\CodeEvidence}{560}
\newcommand{\CodeTallyEvidence}{438}
\newcommand{\CodeRepaired}{62}
\newcommand{\CodeFailed}{5}
\newcommand{\CodeGenCost}{0.10}
\newcommand{\CodeAnswerCost}{0.95}

\newcommand{\ErrN}{214}
\newcommand{\ErrNone}{66}
\newcommand{\ErrNoneNA}{20}
\newcommand{\ErrNoneDropped}{43}
\newcommand{\ErrEv}{148}
\newcommand{\ErrSpec}{51}

\newcommand{\ErrRef}{42}

\newcommand{\ErrSys}{104}
\newcommand{\ErrSpecPct}{49}
\newcommand{\ErrLedgerPct}{20}

\newcommand{\ErrSpecAlt}{26}
\newcommand{\ErrSpecPred}{36}
\newcommand{\ErrSpecOp}{11}

\newcommand{\ErrRefRef}{29}
\newcommand{\ErrRefJudge}{13}
\newcommand{\ErrValA}{39.3}
\newcommand{\ErrValS}{18.1}

\newcommand{\ErrValP}{0.005}
\newcommand{\ErrSpecClear}{25}
\newcommand{\ErrValSClear}{17.0}
\newcommand{\ErrSpecAmb}{26}
\newcommand{\ErrValSAmb}{19.2}

\newcommand{\DisTmFitMin}{$-$2.3}
\newcommand{\DisTmFitMax}{5.5}
\newcommand{\DisTmNofitMin}{8.7}
\newcommand{\DisTmNofitMax}{15.0}

\newcommand{\DisFitLowTm}{82}

\newcommand{\DisFitHighTm}{43}

\newcommand{\DisNofitLowTm}{68}

\newcommand{\DisFitHighGain}{16}
\newcommand{\DisNofitHighGain}{33}
\newcommand{\DisFitLowGain}{2}
\newcommand{\DisNofitLowGain}{$-$5}

\newcommand{\ProbeN}{645}
\newcommand{\ProbeEpisodes}{66}
\newcommand{\ProbeTemplates}{16}

\newcommand{\ProbeEmbEp}{30}
\newcommand{\ProbeSwEp}{36}
\newcommand{\ProbeTally}{93.8}
\newcommand{\ProbeSwTally}{88.0}
\newcommand{\ProbeEmbTally}{99.7}
\newcommand{\ProbeTm}{82.8}

\newcommand{\ProbeCounter}{89.9}
\newcommand{\ProbeSwCounter}{80.2}
\newcommand{\ProbeEmbCounter}{99.7}
\newcommand{\ProbeLong}{49.3}

\newcommand{\ProbeOracle}{94.9}

\newcommand{\ProbeTallyVsTmWL}{+79/$-$8}
\newcommand{\ProbeTallyVsTmP}{$<$0.001}

\newcommand{\ProbeTallyVsCounterWL}{+37/$-$12}
\newcommand{\ProbeTallyVsCounterP}{$<$0.001}

\newcommand{\ProbeAdjTally}{93.8}

\newcommand{\ProbeAdjTm}{81.9}

\newcommand{\ProbeAdjCounter}{89.1}

\newcommand{\ProbeAdjLong}{47.0}

\newcommand{\ProbeWordEmbTmTrig}{100.0}
\newcommand{\ProbeWordEmbTmPlain}{46.7}

\newcommand{\ProbeWordSwTmTrig}{97.2}
\newcommand{\ProbeWordSwTmPlain}{8.3}

\newcommand{\ProbeBinNA}{192}

\newcommand{\ProbeBinNB}{157}

\newcommand{\ProbeBinNC}{147}

\newcommand{\ProbeBinND}{138}

\newcommand{\ProbeEventNonEmpty}{634}
\newcommand{\ProbeEventExact}{558}
\newcommand{\ProbeEventFOne}{96.1}

\newcommand{\ProbeSThreeN}{36}
\newcommand{\ProbeSThreeTally}{30.6}

\newcommand{\ProbeSFourN}{36}
\newcommand{\ProbeSFourTally}{75.0}
\newcommand{\ProbeSFourTm}{86.1}

\newcommand{\ProbeSFourView}{21}
\newcommand{\ProbeSFourWrongSet}{21}
\newcommand{\ProbeSFourWrongSetCorrect}{12}

\newcommand{\ProbeLongCompleteN}{313}
\newcommand{\ProbeLongCompleteEmb}{276}
\newcommand{\ProbeLongCompleteSw}{37}
\newcommand{\ProbeLongFitsEmb}{216}
\newcommand{\ProbeLongVisBOne}{100}
\newcommand{\ProbeLongVisLoadOne}{14}

\newcommand{\ProbeLongVisBTwo}{85}
\newcommand{\ProbeLongVisLoadTwo}{27}

\newcommand{\ProbeLongVisBThree}{80}
\newcommand{\ProbeLongVisLoadThree}{53}

\newcommand{\ProbeLongVisEp}{20}
\newcommand{\ProbeLongVisEmbBinA}{95.8}
\newcommand{\ProbeLongVisEmbBinNA}{120}
\newcommand{\ProbeLongVisEmbBinB}{76.7}
\newcommand{\ProbeLongVisEmbBinNB}{43}
\newcommand{\ProbeLongVisEmbBinC}{90.7}
\newcommand{\ProbeLongVisEmbBinNC}{54}
\newcommand{\ProbeLongVisEmbBinD}{72.9}
\newcommand{\ProbeLongVisEmbBinND}{48}

\newcommand{\ProbeOracleSThree}{22.2}
\newcommand{\ProbeOracleSFour}{97.2}
\newcommand{\ProbeOracleVsTallyWL}{+14/$-$7}
\newcommand{\ProbeOracleVsTallyP}{0.19}
\newcommand{\ProbeOracleSFourWL}{+9/$-$1}
\newcommand{\ProbeOracleSFourP}{0.02}
\newcommand{\ProbeOracleNoSThree}{99.2}
\newcommand{\ProbeCITallyTm}{$+$11.0 [$+$8.8, $+$13.3]}
\newcommand{\ProbeCITallyCounter}{$+$3.9 [$+$1.5, $+$6.5]}
\newcommand{\ProbeCITmLong}{$+$33.5 [$+$27.4, $+$39.6]}
\newcommand{\ProbeCIOracleTally}{$+$1.1 [$+$0.0, $+$2.2]}
\newcommand{\ProbeCIClusters}{66}
\newcommand{\ProbeJudgeReviewed}{47}
\newcommand{\ProbeJudgeLenient}{36}
\newcommand{\ProbeJudged}{3,870}

\newcommand{\RobClusters}{208}
\newcommand{\RobBoot}{10,000}

\newcommand{\RobReexecIdentical}{2,459}
\newcommand{\RobReexecQueries}{1,160}

\newcommand{\RobReexecRecords}{804}
\newcommand{\ProbeCode}{86.2}

\newcommand{\ProbeCodeVsTallyWL}{+13/$-$62}
\newcommand{\ProbeCodeVsTallyP}{$<$0.001}

\newcommand{\ProbeCodeSThree}{19.4}
\newcommand{\ProbeCodeSFour}{69.4}

\newcommand{\ProbeCodeEmb}{91.9}
\newcommand{\ProbeCodeSw}{80.6}
\newcommand{\ProbeCodeSThreeObs}{0}
\newcommand{\ProbeCodeSThreeQuote}{24}
\newcommand{\ProbeCodeSThreeOutcome}{12}
\newcommand{\ProbeCodeRendered}{636}

\newcommand{\ProbeCICodeTally}{$-$7.6 [$-$10.2, $-$5.0]}
\newcommand{\ProbeEightTally}{83.1}
\newcommand{\ProbeEightTm}{70.1}
\newcommand{\ProbeEightCounter}{81.9}
\newcommand{\ProbeEightLong}{54.3}
\newcommand{\ProbeEightOracle}{84.3}

\newcommand{\ProbeEightTallyVsCounterWL}{+49/$-$41}

\newcommand{\ProbeEightWordEmbTmTrig}{56.7}
\newcommand{\ProbeEightWordEmbTmPlain}{26.7}
\newcommand{\ProbeEightWordSwTmTrig}{22.2}
\newcommand{\ProbeEightWordSwTmPlain}{13.9}
\newcommand{\ProbeEightLongVisBOne}{95}

\newcommand{\ProbeEightLongVisBTwo}{100}

\newcommand{\ProbeEightLongVisBThree}{90}
\newcommand{\ProbeEightTallyVisBThree}{75}
\newcommand{\ProbeEightCITallyTm}{$+$13.0 [$+$9.7, $+$16.5]}
\newcommand{\ProbeEightJudgeLenient}{47}
\newcommand{\ProbeEightJudgeReviewed}{103}

\newcommand{\ReaderLoadSizes}{299 / 102 / 37}
\definecolor{readc}{HTML}{75617E}
\definecolor{execc}{HTML}{3D6B59}

\usepackage{authblk}

\makeatletter
\renewcommand{\outauthor}{\parbox{\textwidth}{\centering\@author}}
\makeatother

\title{When Agent Memory Needs to Compute: \\ Evidence Access and Executable Aggregation}
\author{Zerui~Yang\thanks{Correspondence: \texttt{richard.yangpolyu@gmail.com}.}}
\author{Quan~Xue}
\author{Chaorui~Zhang\thanks{Correspondence: \texttt{zhang.chaorui@huawei.com}.}}
\author{Xueyan~Niu}
\author{Xuanyu~Zheng}
\author{Xiyuan~Wang}
\author{Haolin~Ruan}
\author{Guangcheng~Song}
\author{Weixi~Zhang}
\author{Wei~Han}
\author{Bo~Bai}
\affil{Huawei Technologies Ltd.}
\begin{document}
\maketitle

\begin{abstract}
Questions about an agent's history often require aggregating events, which retrieval alone does not resolve. We combine source-preserving evidence access with executable aggregation. Trajectory Memory (TM) retains the original trajectory and constructs a source-addressable reading view at question time. TallyMem augments it with a reusable event ledger, model-generated event specifications and deterministic execution. With a commit-specific adaptation on 776 questions from 99 held-out MINTEval repositories, TallyMem raises Qwen3-32B exact match from \MintTm\% to \MintTally\%, a 13.8-point gain over TM; gains span \MintGainLo--\MintGainHi{} points across three readers. A controlled probe with event sets parsed independently from raw logs shows an 11.0-point gain, from \ProbeTm\% to \ProbeTally\%, without routing. On full AMA-Bench, the routed system yields a smaller 1.3-point gain, from \FullTm\% to \FullGate\%. These results support executable aggregation on tasks that require combining events, with benefits dependent on the task and event specification, the largest residual system-error category in our diagnostic analysis.
\end{abstract}

\section{Introduction}

An agent that edits a repository, navigates a website or operates in a simulated home accumulates a history of
actions and observations. Later questions about that history can ask what happened at one step, how an object's
state changed, or how often an action was taken \citep{zhao2026amabench,he2026memoryarena,xu2026memgym}. Answering them may need the exact action, its result and its position.

\begin{figure*}[t]
\centering
\includegraphics[width=\textwidth]{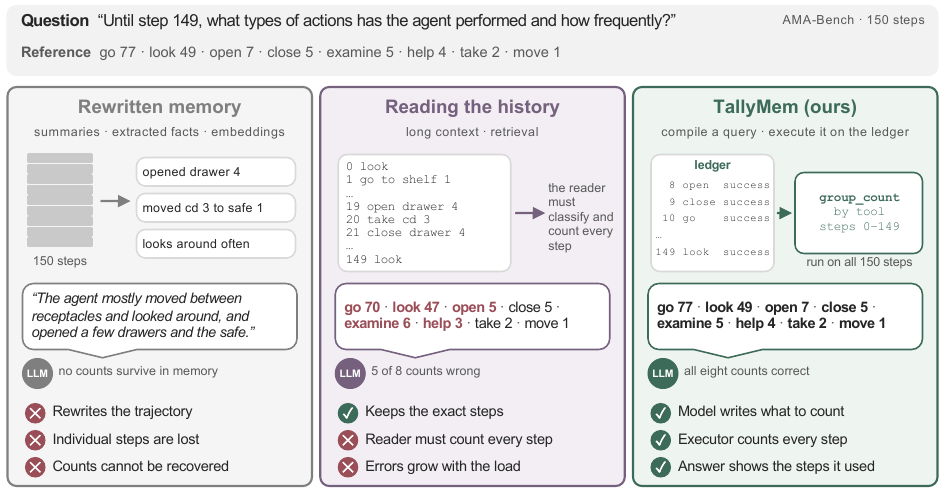}
\caption{Evidence access and event aggregation on an action-frequency question. Left: an illustrative compression example, not a measured output.
Middle: Qwen3-32B's actual long-context answer miscounts five of eight action types.
Right: with \sys's group-count query over all 150 steps, Qwen3-32B reports every count
correctly. Red numbers are wrong.}
\label{fig:case}
\end{figure*}

Such questions place two distinct demands on memory. The first is \emph{access}: the steps that determine the answer
must reach the reader unchanged. The second is \emph{aggregation}: when an answer combines many events, they must be
selected and combined correctly. Common memory designs retrieve evidence from summaries, extracted facts or linked
records \citep{packer2023memgpt,chhikara2025mem0,xu2025amem,gutierrez2024hipporag}. These approaches emphasize evidence access through retention and retrieval; combining the retrieved events remains a separate challenge. Figure~\ref{fig:case} shows why this is not enough:
even a reader given all 150 steps of a trajectory miscounts five of eight action types.

For access, we argue for \emph{deferred evidence selection}. Memory written before the questions are known must guess
what will matter: a summary commits to an interpretation that a later question may contradict, and AMA-Bench
attributes memory errors to lost objective information and causal order \citep{zhao2026amabench}. We therefore keep
the trajectory verbatim and decide what to show only at question time. \textbf{Trajectory Memory (TM)} implements this principle through a lossless, source-addressable archive and a budgeted, order-preserving reading view, built without model calls.

Better access does not remove the second demand. We measure a question's \emph{aggregation load}, the number of events its answer combines. On a controlled probe with event sets parsed from raw logs, long-context reading of fully visible histories answers the same count correctly for \ProbeLongVisBOne\% of trajectories at \ProbeLongVisLoadOne{} events but for \ProbeLongVisBThree\% at \ProbeLongVisLoadThree{} events. On AMA-Bench, a compiler-derived load proxy shows TM falling from \FullBinTmA\% at one to five events to \FullBinTmE\% above a hundred (Figure~\ref{fig:load}). The probe supplies independent event sets; reference-answer-derived loads provide an approximate cross-check (Appendix~\ref{app:robust}). TM's advantage over long-context reading is larger for each of five readers when the history exceeds the context budget, while aggregation errors remain on fully visible probe histories.

\textbf{\sys{}} addresses aggregation by separating deciding what to count from counting. Before any question, it builds
an event ledger from the trajectory. At question time, a model writes an event specification (a scope, an event
predicate and an operation), and a deterministic executor evaluates it over the whole ledger; the result and its
supporting step IDs join TM's evidence. The specification supports counts, lists, first and last events, groupings and timelines over literal filters. This interface makes event selection explicit and aggregation
deterministic, allowing errors in specifying the intended events to be distinguished from errors in combining them.

With a commit-specific adaptation on 99 held-out MINTEval repositories, \sys{} improves exact match by 13.8 points over TM, with gains across three readers. The controlled probe adds 11.0 points without routing. On full AMA-Bench, the routed system gives a smaller gain, from \FullTm\% to \FullGate\% (Table~\ref{tab:full}); gains vary across readers and trajectory splits. Executing specifications also exposes event-selection mistakes as the largest residual system-error category.

Our contributions are:
\begin{enumerate}
  \item A diagnosis of distinct access and aggregation bottlenecks using five readers and controlled event sets.
  \item Complementary mechanisms over a shared history: TM preserves and selects source evidence, while \sys{} compiles event specifications for deterministic aggregation over a reusable ledger.
  \item Held-out repository evaluation and controlled aggregation tests, alongside full-benchmark comparisons and analysis of residual event-specification errors.

\end{enumerate}

\section{Related Work}
\paragraph{Agent memory.}
MemGPT, Mem0 and A-Mem organize experience through summaries or extracted records \citep{packer2023memgpt,chhikara2025mem0,xu2025amem}; HippoRAG retrieves through a knowledge graph \citep{gutierrez2024hipporag}, and MemSearch-o1 constructs query-aligned fragments \citep{zhang2026memsearch}. Evaluation extends from dialogue to agent trajectories \citep{maharana2024locomo,wu2025longmemeval,he2026memoryarena,zhao2026amabench,yang2026memo}. We study evidence access and event aggregation on AMA-Bench and MINTEval \citep{lee2026minteval}.

\paragraph{Executable memory and event specifications.}
Binder generates programs with model calls \citep{cheng2023binder}, ChatDB uses SQL as symbolic memory \citep{hu2023chatdb}, and AMA-Agent executes code over trajectories \citep{zhao2026amabench}. Table~\ref{tab:design_comparison} contrasts their designs with ours. Our archive-and-ledger organization couples explicit event selection with original evidence; known event sets let us separately examine access, selection and aggregation. Appendix~\ref{app:code} tests a matched free-form code alternative.

\begin{table*}[t]
\centering\footnotesize
\setlength{\tabcolsep}{4pt}
\renewcommand{\arraystretch}{1.12}
\begin{tabular}{|p{0.095\textwidth}|p{0.175\textwidth}|p{0.175\textwidth}|p{0.185\textwidth}|p{0.24\textwidth}|}
\hline
Dimension & Binder & ChatDB & AMA-Agent & TM + TallyMem (ours) \\
\hline

Memory & Task input used by generated programs & SQL databases as symbolic memory & Structured trajectory memory with causal links & \textbf{Original archive + reusable event ledger} \\
\hline
Execution & Programs with language-model API calls & Model-generated SQL instructions & Trajectory-based code when retrieval is insufficient & \textbf{Explicit event specification + fixed executor} \\
\hline
Evidence use & Program execution over task input & Database operations and their results & Memory retrieval with code-based processing & \textbf{Aggregates and step IDs alongside original evidence} \\
\hline
Study focus & Program-based question answering & Symbolic memory for multi-hop reasoning & Long-horizon trajectory memory & \textbf{Access, event selection and aggregation examined separately} \\
\hline

\end{tabular}
\caption{Design comparison with executable-query and memory systems \citep{cheng2023binder,hu2023chatdb,zhao2026amabench}. Entries summarize published designs and study objectives; bold highlights our archive--ledger organization and event-level analysis.}
\label{tab:design_comparison}
\end{table*}

\paragraph{Retrieval versus aggregation.}
Available context does not ensure correct use of evidence \citep{liu2024lost}. RULER and Oolong test information aggregation \citep{hsieh2024ruler,bertsch2025oolong}; controlled studies identify instance count as a source of difficulty \citep{dai2026rulefollowing,chen2026multiinstance}. MemTrace finds failures despite retrievable evidence \citep{long2026memtrace}. Our trajectory analysis distinguishes evidence-access failures from errors in selecting and combining events.

\section{Problem Formulation}
\label{sec:problem}

\paragraph{Trajectories and memory.}
Let $H=((i,a_i,o_i))_{i\in\mathcal I_H}$ be an ordered trajectory of actions $a_i$ and observations $o_i$ \citep{yao2022react}. Memory is built from $H$ before questions arrive. Given a later question $q$, it supplies context $C_q$ to a fixed answer model, $\widehat y=F_\theta(q,C_q)$. We compare systems with shared readers and judges; TM's view has budget $B$; computed evidence has its own bound.

\paragraph{Two requirements.}
The first requirement is \emph{access}: the steps that determine the answer must reach $C_q$ unchanged and in order,
so that the reader can use the exact action, its observed result and its position. Many questions also ask for a
quantity over a set of steps,
\begin{equation}
\begin{aligned}
  y^{\star}&=g\big(S^{\star}(q)\big),\\
  S^{\star}(q)&=\{\,i\in\mathcal I_H\cap I_q:\ \phi_q(a_i,o_i)=1\,\},
\end{aligned}
  \label{eq:target}
\end{equation}
where $I_q$ is the step scope named by the question, $\phi_q$ decides which steps are events of the requested kind,
and $g$ counts, groups, orders or lists them. The second requirement is \emph{aggregation}: computing $g$ correctly
over $S^{\star}(q)$, whose size $|S^{\star}(q)|$ we call the question's aggregation load. Access can hold while
aggregation fails: even if $C_q$ contains every step of a 150-step history, a reader asked for action frequencies
must still classify all 150 steps and count each of them once.

TM supplies a source-preserving reading view $R_q$. \sys{} additionally estimates $(I_q,\phi_q,g)$, executes it over a reusable event ledger, and supplies the result $D_q$, giving $C_q=R_q\oplus D_q$. Execution is exact over the selected ledger rows; identifying $S^{\star}(q)$ remains a model decision.

\section{Deferred Evidence Selection: TM}
\label{sec:tm}

What a later question needs is unknown when memory is written: an apparently incidental tool result can decide an
answer many steps later. TM therefore never rewrites the trajectory; it only chooses, once the question is known,
which parts to show \citep{lewis2020retrieval}. Neither memory construction nor retrieval calls a model (Figure~\ref{fig:tm_detail} in
Appendix~\ref{app:implementation}).

\paragraph{Lossless write.}
The archive $\mathcal A=\operatorname{Archive}(H)$ stores every non-empty step verbatim with its original step ID.
A source reference $r=(i,f,[u,v))$ addresses characters $u$ through $v-1$ of field
$f\in\{\mathrm{action},\mathrm{observation}\}$ of step $i$, and every span TM shows satisfies
\begin{equation}
  \operatorname{text}(r)=\mathcal A[i,f][u:v],
  \label{eq:provenance}
\end{equation}
so any retrieved evidence can be traced to the original record.
For editor tool calls \citep{wang2025openhands}, TM links each read or write to a file node keyed by its exact path, confirmed only by the
tool's receipt, so that questions about a file reach its earlier operations; other calls stay in the
archive (Appendix~\ref{app:implementation}).

\paragraph{Reading as anchored selection.}
Given $q$, TM forms an anchor set
\begin{equation}
\begin{aligned}
  A_q &= E_q\cup\operatorname{supp}\!\big(\operatorname{top}_k\, s(\cdot,q)\big),\\
  s(x,q) &= |T(q)\cap T(x)|+\beta(x,q),
\end{aligned}
  \label{eq:anchors}
\end{equation}
where $E_q$ are the steps the question names explicitly, $x$ ranges over the indexed steps and file entities in scope,
$T(\cdot)$ is a term set, $\beta$ rewards explicitly named steps and entities, and $\operatorname{supp}$ maps the
top-scoring items to the steps that support them. TM then admits the steps in scope greedily, named steps first and the
others by their distance to the nearest anchor (Appendix~\ref{app:implementation}),
until the token budget $B$ is spent, and renders them chronologically, marking omitted steps and truncated
fields:
\begin{equation}
\begin{aligned}
 R_q &= \operatorname{Render}_{\mathrm{chron}}\!\left(\operatorname{Greedy}_{B}(I_q;\rho_q)\right),\\
 \widehat y_{\mathrm{TM}} &= F_{\theta}(q,R_q).
\end{aligned}
\label{eq:tmview}
\end{equation}
We use $B=26$k tokens; the remaining constants are in
Appendix~\ref{app:implementation}. The view preserves the source and order of every span it shows but does not
guarantee complete visibility, because steps can be omitted and fields truncated. TM also contains one fixed aggregate,
a keyword-triggered action counter, which we examine in Section~\ref{sec:probe}.

\section{Diagnosing Access and Aggregation Bottlenecks}
\label{sec:diagnosis}

We first diagnose TM on AMA-Bench \citep{zhao2026amabench}: 2,496 questions about 208 real trajectories from six
domains (setup in Section~\ref{sec:setup}).

\paragraph{Aggregation load: how many events must be combined?}
History length measures how much text is available; aggregation load measures how many event occurrences contribute to an answer. In a 150-step history, a question about one particular opening may need only one event, whereas asking for the frequency of every action requires all 150. Two equally long histories can therefore impose very different aggregation demands.

Ideally, load is $|S^{\star}(q)|$ in Equation~\ref{eq:target}. AMA-Bench provides reference answers but not the required event sets, so we estimate it from executable queries (Section~\ref{sec:compile}). If $S_j$ holds the events selected for quantity $j$, the load is
\begin{equation}
 \ell(q,H)=\sum_{j\in J}|S_j|,
 \label{eq:load}
\end{equation}
where $J$ contains the sub-queries that pass the grounding checks. Separate counts of seven openings and five closings have $\ell=12$. An event used in two requested quantities counts twice. This is a \emph{compiler-derived proxy}, not an independent difficulty annotation: an incorrect filter can change the measured load, and a question without a surviving query has undefined load. Reference-answer-derived loads parse counts and cited steps as a further approximation (Appendix~\ref{app:robust}); Section~\ref{sec:probe} instead uses event sets parsed independently from raw logs. Compiler-defined bins are descriptive diagnostics, not independent difficulty strata.

Figure~\ref{fig:load} (left) bins the \FullEvidN{} questions with a load. TM falls from \FullBinTmA\% at one to five
matched steps to \FullBinTmE\% above a hundred; the decline also holds within the two domains rich in high-load samples.

\begin{figure}[t]
\centering
\begin{tikzpicture}
\pgfplotsset{loadpanel/.style={scale only axis,width=3.05cm,height=2.9cm,ymin=0,ymax=100,ytick={0,25,50,75,100},
  grid=major,grid style={gray!20},tick label style={font=\scriptsize},label style={font=\small},
  title style={font=\small,yshift=-0.8ex},x tick label style={font=\tiny},xlabel style={yshift=0.6ex}}}
\begin{axis}[loadpanel,name=ama,title={AMA-Bench},ylabel={accuracy (\%)},xlabel={matched steps},xtick={0,1,2,3,4},
  xticklabels={1--5,6--15,16--40,41--100,$>$100}]
\addplot[color=readc,mark=*,thick] table[x=x,y=tm]{figures/load_full.dat};
\end{axis}
\begin{axis}[loadpanel,at={(ama.south east)},anchor=south west,xshift=0.25cm,yticklabels={},title={Probe},
  xlabel={logged events},xtick={0,1,2,3},xticklabels={1--5,6--15,16--40,$>$40},
  legend style={font=\scriptsize,draw=none,fill=none,at={(0.03,0.03)},anchor=south west},legend cell align=left]
\addplot[color=readc,mark=*,thick] table[x=x,y=tm]{figures/probe_load.dat};
\addlegendentry{TM}
\addplot[color=execc,mark=*,thick] table[x=x,y=tally]{figures/probe_load.dat};
\addlegendentry{\sys}
\addplot[color=gray,mark=square*,mark size=1.5pt,thick,dashed] table[x=x,y=counter]{figures/probe_load.dat};
\addlegendentry{fixed counter}
\end{axis}
\end{tikzpicture}
\caption{Compiler-defined load diagnostic and log-defined aggregation probe. Left: TM (Qwen3-32B) on the \FullEvidN{} AMA-Bench questions with
query evidence, by the steps their compiled queries match (\FullBinNA, \FullBinNB, \FullBinNC, \FullBinND{} and
\FullBinNE{} questions). Right: the probe, by the size of the event set parsed from the raw log (\ProbeBinNA,
\ProbeBinNB, \ProbeBinNC{} and \ProbeBinND{} questions); the fixed counter groups steps by tool, and TM's curve includes
its counter on ``how many times'' questions.}
\label{fig:load}
\end{figure}

\begin{figure}[t]
\centering
\includegraphics[width=0.93\columnwidth]{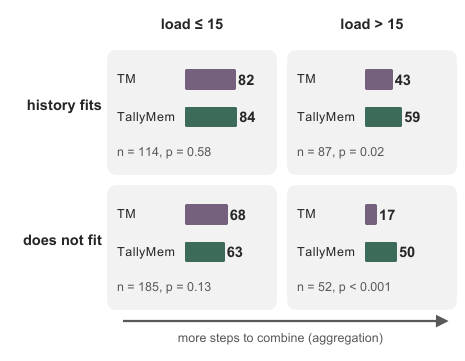}
\caption{Two constraints on trajectory answering. Qwen3-32B on 438 rule-selected diagnostic questions with a
measured load, with computed evidence always attached. ``History fits'' means at most 24k raw-history tokens,
not full visibility of every TM field. Cells show accuracy (\%), sample size and paired sign-test $p$.}
\label{fig:twoaxis}
\end{figure}

\paragraph{Access and aggregation fail separately.}
Figure~\ref{fig:twoaxis} crosses nominal context fit with load. When the history fits, TM still falls from
\DisFitLowTm\% at low load to \DisFitHighTm\% at high load; when it does not fit, even low-load questions drop to
\DisNofitLowTm\%. The two interventions act on different cells. Across the five readers, TM's gain over
long-context reading lies between \DisTmFitMin{} and \DisTmFitMax{} points on histories that fit and between
\DisTmNofitMin{} and \DisTmNofitMax{} points on histories that do not, larger for every reader
(Appendix~\ref{app:readers}). Execution instead helps at high load in both strata (+\DisFitHighGain{} and
+\DisNofitHighGain{} points) and not at low load (+\DisFitLowGain{} and \DisNofitLowGain{} points), where reading is already accurate and a
misspecified query can mislead it \citep{shi2023large}. Fully visible histories, larger or
reasoning-oriented readers and prompts that enumerate steps show the same decline (Appendices~\ref{app:visibility}
and~\ref{app:readers}).

\section{Separating Specification from Execution: TallyMem}
\label{sec:method}

\begin{figure*}[t]
\centering
\includegraphics[width=\textwidth]{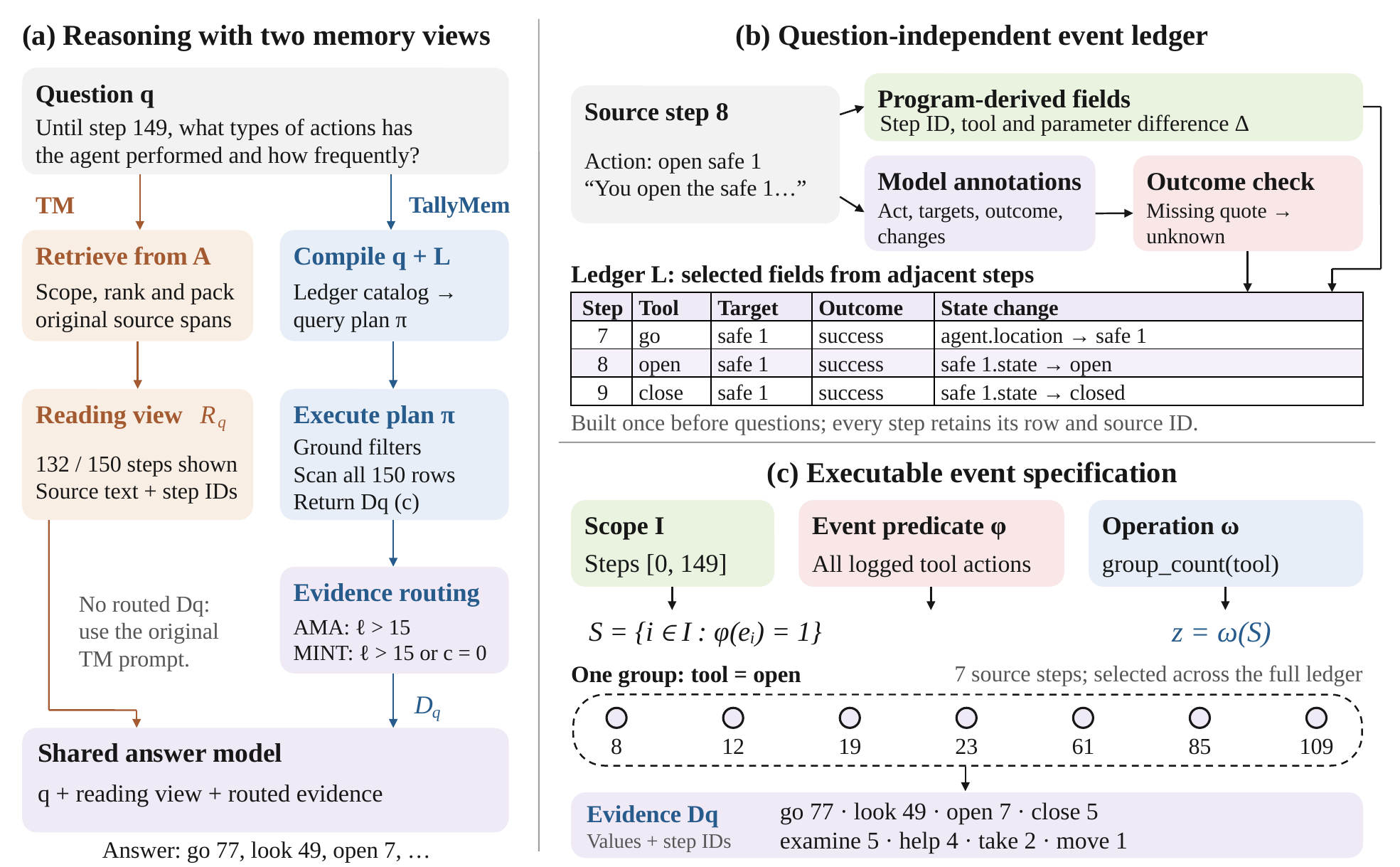}
\caption{TM and \sys{} through a shared running example. (a)~TM retrieves source text from archive $\mathcal A$;
\sys{} adds computed evidence to the same reader. (b)~The AMA ledger combines program-derived
fields with model annotations before questions arrive. The outcome check verifies quote occurrence, not semantic
correctness; MINTEval uses a programmatic ledger. (c)~The model specifies scope, event predicate and operation;
the executor selects and groups all 150 rows. The seven circles are the \texttt{open} group; its count and
step IDs enter $D_q$.}
\label{fig:overview}
\end{figure*}

\sys{} takes aggregation away from the reader. A model writes the specification $(I_q,\phi_q,g)$ of
Equation~\ref{eq:target} as a plan in a small typed query language, and an executor evaluates the plan exactly over an
event ledger of the whole trajectory. The model therefore decides \emph{what} to count but never does the counting \citep{lyu2023faithful}.
TM is unchanged, and the result is added to its reading view (Figure~\ref{fig:overview}).

\paragraph{The ledger: a grounded relation over steps.}
At write time \sys{} turns the archive into a relation $L=\Phi(\mathcal A)=(e_i)_{i\in\mathcal I_H}$ with one row per
logged step, built once per trajectory and shared by all its questions. A row has the schema
\[
\begin{aligned}
 e_i=(\,&i,\ \mathrm{tool},\ \mathrm{act},\ \mathrm{action},\\
        &\mathrm{targets},\ \mathrm{outcome},\ \mathrm{changes},\ \Delta\,),
\end{aligned}
\]
listing the tool, normalized action, raw action, targets, outcome, state changes and parameter difference $\Delta$. A
program extracts the mechanical fields; a model annotates targets, outcomes and changes from log chunks without seeing
questions \citep{arora2023language}. Outcomes need a quote found in the observation, changes a successful step, and uncertain annotations never remove a row (Appendix~\ref{app:implementation}).

\paragraph{A typed query language.}
\label{sec:compile}
A plan is either \emph{not applicable} or a list $\pi=(p_1,\ldots,p_k)$ of at most four sub-queries, one per quantity
the question asks for. Each sub-query $p=(I,\phi,\omega)$ has a step scope $I=[\mathit{lo},\mathit{hi}]$ with possibly open bounds,
a predicate and an operation. The predicate is a conjunction over fields of disjunctions of literal terms,
\begin{equation}
  \phi(e)=\bigwedge_{f\in F}\ \bigvee_{t\in T_f}\big[\,t\sqsubseteq e.f\,\big],
  \label{eq:predicate}
\end{equation}
where $F$ is a subset of the tool, action, target, outcome, change and $\Delta$ fields and $t\sqsubseteq e.f$ denotes a
case-insensitive substring match (equality for the outcome; tool terms match the tool or act). With $S=\{i\in\mathcal I_H\cap I:\phi(e_i)=1\}$ and $\kappa_f(i)$ the set of
values of field $f$ in row $i$ (one tool per step, possibly several targets), the operations are
\begin{equation}
\begin{aligned}
 \texttt{count}&:\ |S|,\\
 \texttt{list}&:\ (e_i)_{i\in S},\\
 \texttt{first}/\texttt{last}&:\ e_{\min S}\ /\ e_{\max S},\\
 \texttt{group}_f&:\ g\mapsto\textstyle\sum_{i\in S}\mathbf 1\{g\in\kappa_f(i)\},\\
 \texttt{timeline}&:\ \tau\mapsto(e_i)_{i\in S_\tau},
\end{aligned}
\label{eq:operations}
\end{equation}
where $S_\tau=\{i\in S:\tau\in\kappa_{\mathrm{target}}(i)\}$ and a timeline reports each row's step, tool, outcome and
changes. Rows are always taken in step order, and $\texttt{group}_f$ (group-count) runs over the tool, target or
outcome field. The language provides a fixed set of event-selection and aggregation operators. Each plan can be checked against the ledger and evaluated in time linear in $|L|$. The supported operations and their scope are detailed in Appendix~\ref{app:implementation}.

\paragraph{Compilation.}
Besides the shared answer model, \sys{} makes one question-time model call for compilation,
$\pi=\operatorname{Compile}(q,\operatorname{catalog}(L))$. The catalog summarizes tools, targets and change properties, so that filters use the ledger's own vocabulary (Appendix~\ref{app:implementation}). The scope comes only from explicit
bounds in the question (``until step 149'' becomes $[0,149]$) and is otherwise the whole trajectory. Questions
about explanations, causes, advice or the content of a single observation receive TM's view alone.

\paragraph{Grounded execution.}
Before a sub-query runs, the executor checks it against $L$: every filter term must occur in the corresponding field
somewhere in the log, and the match set $S$ must be non-empty. A sub-query that fails either check contributes nothing,
because an unmatched term or an empty match signals a specification the log cannot support, not a verified zero; its
valid siblings survive. The one exception is a list or timeline query with no match, which is retried once with only its
target and action filters and is then marked as relaxed. For the surviving set $J$, the executor computes
$z_j=\omega_j(S_j)$ by Equation~\ref{eq:operations} and renders the results in plan order,
$D_q=\operatorname{Render}_{M}((p_j,z_j)_{j\in J})$, within a budget of $M=6{,}000$ characters. Each result states its
operation, filters and number of matched steps, followed by the computed values and the step IDs that support them
(Figure~\ref{fig:example}).

\paragraph{Answering.}
The shared reader returns
\begin{equation}
 \widehat y_{\gamma}=
 \begin{cases}
 F_{\theta}(q,R_q\oplus D_q), & D_q\ne\varnothing,\ \gamma=1,\\
 F_{\theta}(q,R_q), & \text{otherwise}.
 \end{cases}
 \label{eq:augmentation}
\end{equation}
We add a router \citep{jeong2024adaptive} to decide whether to attach computed evidence ($\gamma$; Appendix~\ref{app:gate}).
Without a block the prompt is exactly TM's, and we reuse the TM answer.

\paragraph{Guarantee and boundary.}
The executor aggregates every selected row, including rows omitted from $R_q$. If the ledger and specification
represent the intended events, it computes $g(S^{\star}(q))$ exactly; otherwise the error lies in one of them, and the
returned conditions and step IDs show which (Section~\ref{sec:analysis}; worked example in Appendix~\ref{app:examples}).

\section{Evaluation}
\label{sec:results}

\begin{table}[!t]
\centering\small
\setlength{\tabcolsep}{4pt}
\begin{tabular}{@{}lrrr@{}}
\toprule
System & Qwen3-32B & gpt-oss-120b & Qwen3-8B \\
\midrule
\multicolumn{4}{@{}l}{\emph{AMA-Bench: accuracy (\%), 2,496 questions}} \\
Long context & \FullLong & \FullOssLong & \FullEightLong \\
BM25 & \FullBlBm & -- & -- \\
Dense RAG & \FullBlDense & -- & -- \\
Rolling summary & \FullBlRoll & -- & -- \\
Mem0 & \FullBlMemzero & -- & -- \\
A-Mem & \FullBlAmem & -- & -- \\
\midrule
TM & \FullTm & \FullOssTm & \FullEightTm \\
\sys & \textbf{\FullGate} & \textbf{\FullOssGate} & \textbf{\FullEightGate} \\
\midrule
\multicolumn{4}{@{}l}{\emph{MINTEval GitHub: exact match (\%), \MintN{} questions}} \\
Long context & \MintLong & \MintOssLong & \MintEightLong \\
TM & \MintTm & \MintOssTm & \MintEightTm \\
\sys & \textbf{\MintTally} & \textbf{\MintOssTally} & \textbf{\MintEightTally} \\
\bottomrule
\end{tabular}
\caption{Main results. Each column compares systems within one setting; external memory systems run in the same
deployment with the Qwen3-32B reader. MINTEval uses 99 held-out repositories; its long-context row and its
gpt-oss-120b and Qwen3-8B entries were run after the pre-registered test. Paired tests and configurations appear in
Appendices~\ref{app:full} and~\ref{app:mint}.}
\label{tab:full}
\end{table}
\subsection{Setup}
\label{sec:setup}

We evaluate long-context reading, TM and \sys{} on all 2,496 AMA-Bench questions with its official model-judge prompt,
and on 776 questions from 99 held-out repositories of MINTEval's GitHub commit-history subset \citep{lee2026minteval}
with exact match. Aggregation diagnostics compare five readers on 630 common questions, and a controlled probe adds
\ProbeN{} generated questions with known event sets (Section~\ref{sec:probe}). Each comparison uses one answer model,
and the AMA set includes development trajectories. Full evaluations use the frozen principal system; diagnostics attach computed evidence unconditionally. Paired tests are in appendix tables; splits, deployments, selection rules and answer reuse are in Appendices~\ref{app:protocol}
and~\ref{app:gate}.

\subsection{AMA-Bench: Aggregation Gains and Overall Performance}

Full-set gains over TM are modest. With Qwen3-32B \citep{yang2025qwen3}, TM reaches \FullTm\% and \sys{} \FullGate\%, a 1.3-point gain; gpt-oss-120b \citep{agarwal2025gpt} reaches \FullOssTm\% and \FullOssGate\% (+0.8), and Qwen3-8B \citep{yang2025qwen3} reaches \FullEightTm\% and \FullEightGate\% (Table~\ref{tab:full}). Qwen3.6-27B \citep{yang2025qwen3} improves overall (Table~\ref{tab:threshold}), but its held-out-half gain is only 0.1 points, with a 95\% trajectory-bootstrap interval of [$-$0.7, $+$1.0]. Without routing, Qwen3-32B reaches 70.4\% (+79/$-$63 versus TM, $p=0.21$); the overall gain is not significant under the paired sign test. Appendix~\ref{app:gains} reports gains, intervals and trajectory splits.

On the \FullHighN{} questions where the router attaches nonempty evidence (compiled load above 15), Qwen3-32B rises from \FullHighTm\% to \FullHighEv\% (+47/$-$14). These are the same 33 net additional correct answers as in the full-set routed comparison. The subset explains that gain but is selected by the method, rather than independently defining aggregation difficulty. TM also exceeds the external memory systems in the same Qwen3-32B deployment, of which BM25 is strongest at \FullBlBest\%.
Removing only the counts of TM's keyword-triggered counter (\NoCtrTrig{} questions) leaves TM at \NoCtrTm\%,
still \NoCtrGain{} points above long context (Appendix~\ref{app:gains}). TM achieves state-of-the-art performance relative to the previously reported AMA-Bench methods compared here (Appendix~\ref{app:reported_methods}).

\subsection{Held-Out Repository Histories: Cross-Dataset Evaluation}

MINTEval tests the ledger-and-execution approach on commit histories with held-out repositories \citep{lee2026minteval}. We adapt the
ledger and compiler prompt to commits: a program extracts file additions,
deletions, renames and added assignment lines from diffs. The adaptation was developed on 25 repositories and
evaluated once on 99 held-out ones.

\sys{} raises exact match from \MintTm\% for TM to \MintTally\%, a 13.8-point gain, and exceeds long-context reading at \MintLong\% by 9.8 points
(Table~\ref{tab:full}), with the strongest gains in file order and value history. Answering from the same evidence,
gpt-oss-120b and Qwen3-8B show the same ordering: \MintOssLong\% and \MintEightLong\% with long context, \MintOssTm\%
and \MintEightTm\% with TM, and \MintOssTally\% and \MintEightTally\% with \sys{}. The gain comes from the computation path,
which also reaches commits outside TM's reading view: with every reader TM alone trails long context, because commit questions can lose evidence to the retrieval budget even when they
involve few events (Appendix~\ref{app:mint}).

\subsection{Controlled Probe: Testing the Aggregation Mechanism}
\label{sec:probe}

To isolate aggregation load and event selection under controlled conditions, we generated \ProbeN{} questions from \ProbeTemplates{} templates over all \ProbeEpisodes{} embodied and software
trajectories of AMA-Bench (Appendix~\ref{app:probe}). Answers and event sets $S^{\star}$ are parsed from the raw action
and observation strings, without the ledger, the compiler or a model. The templates cover six operations, three step bounds of one trajectory and two
wordings of the same count. Templates, bounds and analyses were fixed before any answer was generated. This probe tests the aggregation mechanism; the full AMA and held-out repository evaluations establish performance on benchmark questions. Appendix~\ref{app:probe} repeats the probe with Qwen3-8B.

Without routing, Qwen3-32B reaches \ProbeTally\% with \sys{} against \ProbeTm\% with TM, a gain of \ProbeCITallyTm{} points (95\% interval from resampling trajectories). Much of TM's deficit comes from its keyword-triggered counter:
at the same bound, TM answers \ProbeWordEmbTmTrig\% of the embodied and \ProbeWordSwTmTrig\% of the software counts
correctly when they ask ``how many times'', but \ProbeWordEmbTmPlain\% and \ProbeWordSwTmPlain\% when they ask ``how
often'', whereas \sys{} is unaffected. A fixed group-by-tool counter run by the same executor reaches \ProbeCounter\%;
it falls short in software trajectories, whose events are delimited by paths, commands or sub-tools. As load grows, TM declines while \sys{} does not (Figure~\ref{fig:load}, right). Long context reaches only
\ProbeLong\%, partly because truncation cuts every software trajectory; on \ProbeLongVisEp{} fully visible embodied
trajectories, its accuracy on the same count falls from \ProbeLongVisBOne\% to \ProbeLongVisBThree\% as mean load
grows from \ProbeLongVisLoadOne{} to \ProbeLongVisLoadThree{} events.

\section{From Miscounting to Misspecification}
\label{sec:analysis}

Execution removes miscounting where the specification is right: on the probe, a correct plan written from each
template's definition, run by the same executor and reader, answers \ProbeOracleNoSThree\% of the questions from
templates whose events the ledger records (Appendix~\ref{app:probe}). We then analysed every error \sys{} makes on the \CodeN{} diagnostic questions with Qwen3-32B \citep{pan2023logic}
(Appendix~\ref{app:taxonomy}).

\begin{table}[t]
\centering\small
\begin{tabular}{lrr}
\toprule
Category & Errors & Other readers (\%) \\
\midrule
Specification & 51 & 18.1 \\
Ledger & 21 & 38.1 \\
Answering & 21 & 39.3 \\
Partial coverage & 11 & 50.0 \\
Reference or judge & 42 & 17.3 \\
Unclear & 2 & 0.0 \\

\bottomrule
\end{tabular}
\caption{The \ErrEv{} errors \sys{} makes with computed evidence (Qwen3-32B, \CodeN{} diagnostic questions;
Appendix~\ref{app:taxonomy}). Last column: accuracy (\%) of the other four readers given the same evidence.}
\label{tab:errors}
\end{table}

\paragraph{Where the errors are.}
Of the \ErrN{} errors, \ErrNone{} receive no computed evidence and are TM's own. We labelled the other \ErrEv{} with
a decision procedure fixed before inspection: a reference or judging error if the answer agrees with the log;
otherwise an answering error if the evidence suffices, partial coverage if the error lies outside the evidence, a
specification error if the plan does not express the referenced quantity, and a ledger error if it does but the rows
are wrong. None arises in execution or in truncating the evidence (Table~\ref{tab:errors}), and an independently
written matcher reproduces every stored result (Appendix~\ref{app:gains}). Specification is the largest category: \ErrSpec{} errors, \ErrSpecPct\% of the \ErrSys{} that remain after
excluding reference, judging and unclear cases, against \ErrLedgerPct\% each for the ledger and the final answer. Another \ErrRef{} answers agree with the log
but are rejected because the reference is wrong (\ErrRefRef) \citep{northcutt2021confident} or the judge errs (\ErrRefJudge). Other readers given the same
evidence answer \ErrValA\% of the answering errors but only \ErrValS\% of the specification errors, so the latter are
systematic.

\paragraph{What misspecification looks like.}
Most specification errors select the wrong events (\ErrSpecPred{} of \ErrSpec). For questions about actions on an
object, the reference counts only opening and closing it, whereas the plan lists every step that names
it; a literal filter for \texttt{drawer 1} also matches \texttt{drawer 10}; ``between Step 6 and Step 9'' is read
inclusively. Others choose the wrong operation (\ErrSpecOp). In \ErrSpecAlt{} cases the plan's reading is defensible and the reference adopts
another. On the probe, where event sets are parsed from the log, \ProbeSFourView{} of \ProbeSFourN{} plans for a question
that names the create, replace and insert commands also count \texttt{view}, and the correct plan raises them from \ProbeSFourTally\% to \ProbeOracleSFour\%; non-zero
exit codes, which the ledger does not record, are approximated by the failure outcome. Because each result states its
filters and step IDs, such errors can be checked against the log, whereas a reader's miscount leaves no
trace (Appendix~\ref{app:errors}).

\paragraph{More freedom moves the same error.}
Three controls point the same way: five sampled plans repeat the same choice \citep{chen2023universal} (Appendix~\ref{app:ablation}); TM's
fixed counter fails when the wording differs (Section~\ref{sec:probe}), so a fixed aggregate also fails on
specification; and free-form Python over the same ledger and log reaches \CodeAll\% (TM \CodeTm\%, \sys{}
\CodeTally\%) \citep{wang2024executable} (Appendix~\ref{app:code}); on the probe it reaches \ProbeCode\% against \sys's \ProbeTally\%. Detailed code comparisons and error examples appear in Appendices~\ref{app:code} and~\ref{app:errors}.

\section{Conclusion}
TM preserves source evidence; TallyMem makes event selection explicit and executes aggregation. The controlled probe supports the aggregation mechanism, and the adapted MINTEval evaluation shows gains on held-out repository histories. Full AMA-Bench gains are modest and depend on routing and the reader. Executable aggregation therefore offers a task-dependent benefit, with event specification the main residual challenge.

\clearpage
\section*{Limitations}
Our evidence covers AMA-Bench, a controlled probe built from its embodied and software trajectories, and the GitHub
subset of MINTEval, not all agent or dialogue memory tasks. Gains vary across question types and domains (Appendices~\ref{app:readers} and~\ref{app:domain}) and trajectory splits (Appendix~\ref{app:gains}). Appendix~\ref{app:gains} discusses judging, reference answers, method development and the single-annotator error analysis. The ledger and the query language cannot express every event predicate, and on MINTEval TM alone trails long
context.

\section*{Ethics Statement}
We use public agent-trajectory and repository-history benchmarks and collect no new personal data.
Trajectory content is treated as untrusted data. The ledger builder does not execute commands contained in a
trajectory; compiled queries select operations from a fixed executor rather than running arbitrary model-written code.
The free-form code control runs model-written programs only in an operating-system sandbox without network or file access.

\section*{Use of AI assistance.}
AI tools were used only for language polishing and assistance with writing code. The research ideas were developed by the authors.

\bibliography{references}
\clearpage
\appendix
\setcounter{dbltopnumber}{3}\renewcommand{\dbltopfraction}{0.9}\renewcommand{\dblfloatpagefraction}{0.8}
\setcounter{topnumber}{3}\renewcommand{\topfraction}{0.9}\renewcommand{\floatpagefraction}{0.8}
\setcounter{totalnumber}{6}\renewcommand{\textfraction}{0.08}
\section{Evaluation Protocol}
\label{app:protocol}
\paragraph{Questions and freezing.}
AMA trajectories are assigned by hash to two halves, and the method was developed on one of them; all reported
results pool both halves. The rule-selected set contains 670 questions; the common answered set used for the
five-reader comparison contains 630. The principal configuration and its per-question evidence were frozen after
development. A later compiler refinement adds two compile rules chosen after inspecting the principal compiler's errors. The full AMA run uses
frozen principal evidence and compiles every question; nonempty evidence is produced for \FullEvidN{} questions,
including \FullEvidOtherN{} outside the rule-defined families. Family membership is based on question text alone.

\paragraph{Main evaluation configuration.}
The method was developed on half of the AMA trajectories, and the principal compiler configuration was then
frozen; all AMA results pool both halves. Five-reader diagnostics use 630 common questions from pre-specified aggregation families;
full-benchmark evaluation covers all 2,496 questions with Qwen3-32B and gpt-oss-120b, and with Qwen3.6-27B and
Qwen3-8B on our local server. These diagnostic and full runs
use different deployments, so their absolute accuracies are not interchangeable. When a variant produces the
identical answer prompt, we reuse that answer rather than introduce another sampling difference.
The full evaluations use the router defined in Appendix~\ref{app:gate}; diagnostic analyses attach every nonempty computed-evidence block.

\paragraph{Models and deployment.}
The diagnostic Qwen3-32B reader uses thinking, temperature 0, a 12,288-token output limit and a 40,960-token context
on DeepInfra. Additional readers are Qwen3-235B-A22B-Instruct-2507 (no thinking), gpt-oss-120b (reasoning),
Qwen3.6-27B and Qwen3-8B (both thinking, local vLLM). The full Qwen3 run uses local bf16 weights rather than the diagnostic fp8 deployment;
the full gpt-oss run uses DeepInfra. Method comparisons are within a setting. Contexts use exact Qwen3 token counts,
so readers in a diagnostic comparison see the same text. Long-context controls read the original trajectory within
the answer budget; over-budget histories are truncated. TM retrieval is question-specific.

\paragraph{Judge and answer reuse.}
Annotation, compilation and AMA judging use Qwen3-32B without thinking. AMA uses the official judge prompt with
strict yes/no extraction; the judge model is shared across readers. 
A second judge, gpt-oss-120b with the same prompt, agrees with the official judge on 86\% of TM and 88\% of \sys{} judgements (599 questions). It is stricter, but the comparison is unchanged: 52.6\% vs.\ 56.1\% (+47/$-$26, $p=0.02$).
\unskip
When computed evidence is empty, the prompt is the TM prompt and we reuse its answer. Identical evidence in a later
variant also reuses the corresponding answer. Otherwise the prompt is answered afresh. Paired sign tests count
per-question wins and losses.

\paragraph{Cross-deployment check.}
On \FullAgreeN{} common TM questions, the two Qwen3 deployments agree on \FullAgreePct\% of judgments
(\FullAgreeServer\% versus \FullAgreeDeepinfra\% accuracy). This check does not make the deployments interchangeable.

\section{Rule-Based Question Families}
\label{app:rules}
A question belongs to a family when its lower-cased text matches:
\begin{itemize}
  \item count/frequency: \texttt{how many}, \texttt{number of}, \texttt{count(s|ed)}, \texttt{how often/frequently},
  \texttt{frequen};
  \item first/last change: \texttt{first|last|final|latest|earliest} followed within 80 characters by a change or
  outcome word;
  \item state tracing: \texttt{state change}, \texttt{what changes}, \texttt{changes occurred},
  \texttt{(state|location) of \ldots\ at step}, \texttt{changes history}, \texttt{throughout the trajectory};
  \item action enumeration: \texttt{(which|what) (types of) actions}, \texttt{actions \ldots\ performed}.
\end{itemize}

\section{Implementation Details and Variants}
\label{app:implementation}

\begin{figure*}[t]
\centering
\includegraphics[width=\textwidth]{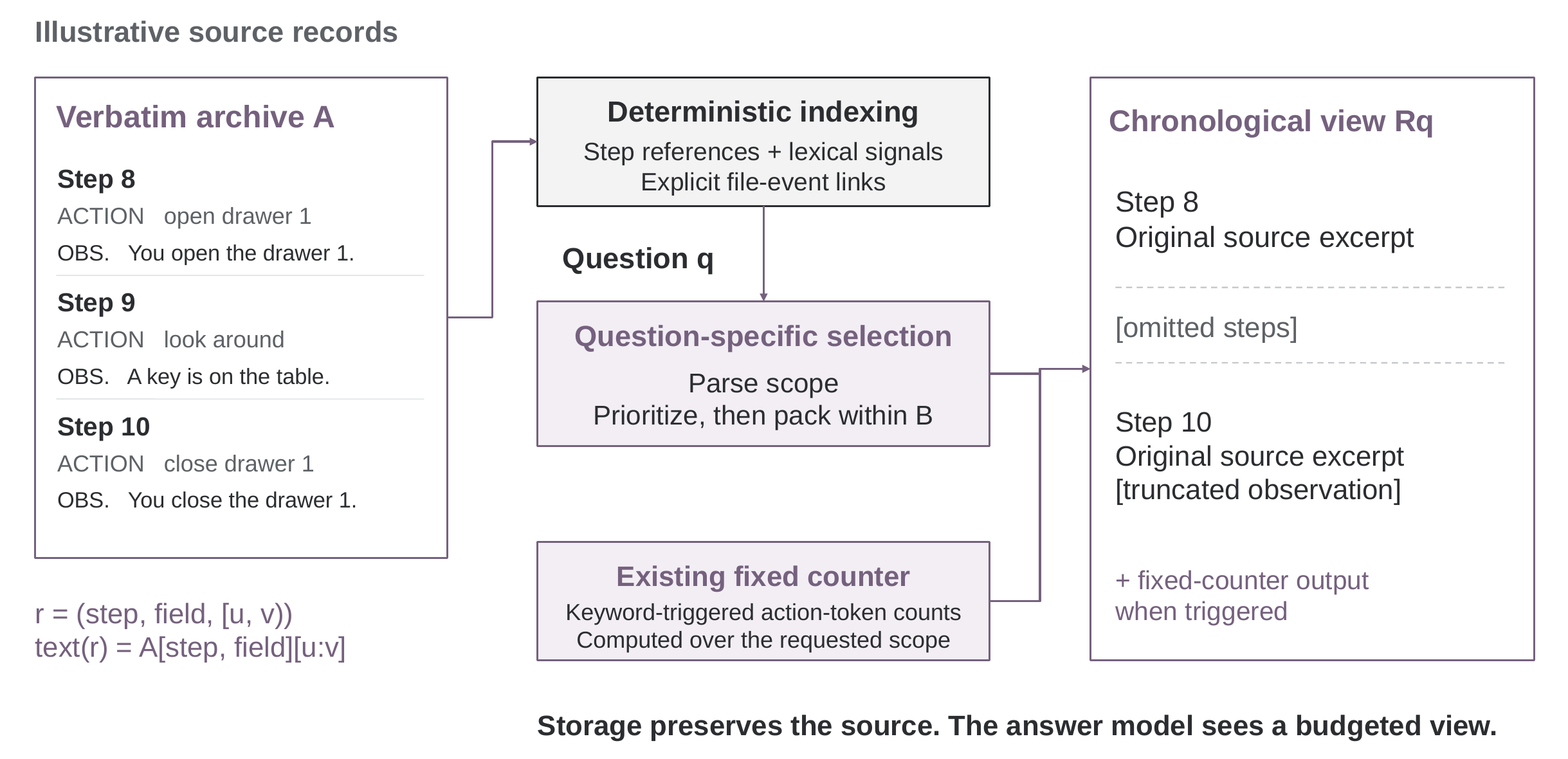}
\caption{TM's archive and reading view (illustrative records). Deterministic indexes guide budgeted selection;
the displayed spans retain their original order and provenance. Storing every step does not guarantee that every
step or field reaches the reader. The keyword-triggered fixed counter is part of TM and remains present in the
\sys{} comparison.}
\label{fig:tm_detail}
\end{figure*}

\paragraph{File-event index.}
TM derives a bipartite graph $G_H=(V_{\mathrm{file}}\cup V_{\mathrm{event}},E)$ from supported editor calls. It parses each \texttt{str\_replace\_editor} action, creates a file node keyed by the exact absolute path, and links the step's event node to that file. The command determines the edge type: \texttt{view} is a read; \texttt{create}, \texttt{insert} and \texttt{str\_replace} are writes. An operation is confirmed only when its observation contains the expected tool-specific leading receipt; otherwise it remains an attempt. Each event stores source spans for its path, command and observation. Thus a question about a file can retrieve its earlier reads and edits through a shared file node, even when the steps are far apart. Unsupported calls remain in the raw archive. The graph neither merges ambiguous basenames nor infers causal links or unseen file states.

\paragraph{Selection order.}
Every step in scope receives the priority
\begin{equation}
\begin{aligned}
  \rho_q(i) &= \big(\mathbf 1\{i\notin E_q\},\ d_q(i),\ i\big),\\
  d_q(i) &= \textstyle\min_{a\in A_q}|i-a|,
\end{aligned}
  \label{eq:priority}
\end{equation}
ordered lexicographically, so named steps come first and the remaining steps follow by their distance to the nearest
anchor of Equation~\ref{eq:anchors}. TM admits steps greedily in this order, skipping any step that no longer fits,
until the budget $B$ is spent; steps linked in $G_H$ to files named in $q$ are admitted from a reserved share of the
budget, and $\operatorname{Greedy}_B$ in Equation~\ref{eq:tmview} denotes this procedure. Fields of steps within
distance $w=2$ of an anchor keep a wider character cap than distant ones.

\paragraph{TM selection constants.}
The score in Equation~\ref{eq:anchors} is the number of shared terms, plus 8 for a step the question names and 12 for
an entity whose label appears in the question; step references are removed from the question before term matching.
The anchor set adds the steps supporting the eight highest-ranked items, or the four best-scoring steps when the
question names nothing. Steps within $w=2$ of an anchor keep 4,000 characters per field and other steps 1,200.
Thirty percent of the budget holds index records rendered next to the steps, and up to eight steps linked in the
file-event graph to files named in the question are admitted from 20\% of the budget that remains after the named
steps and their neighbours.

\paragraph{Ledger fields and validation.}
The mechanical tool is the action's leading identifier, refined by a JSON \texttt{command} argument. Parameter
differences compare actions of the same tool and path. Chunked annotation is question-independent and supplies
a normalized action type, targets, an outcome (success, failure, no effect or unknown), an observation quote, and proposed object.property
changes. The initial validator requires a normalized supporting quote of at least four characters to occur in the
observation. Missing annotations remain unknown, and every step retains a ledger row. Changes require a successful
outcome; a separate \texttt{verified} flag records whether their new value occurs in the observation or action text.
A false flag does not remove the change, and the executor does not filter changes by this flag.

\paragraph{Ledger rows and validation.}
\begin{itemize}
  \item Accepted: \texttt{open drawer 1} is annotated as a success, quoting ``You open the drawer 1. The drawer 1 is
  open'', with the change \texttt{drawer 1.state$\to$open} (verified).
  \item Rejected: an annotation quoting ``You see:...'' is not a substring of its observation, so its outcome becomes
  unknown.
\end{itemize}

\paragraph{Quote recovery.}
The frozen computation path retries quote matching after a looser normalization when a row was rejected for quote
formatting. Recovery requires at least eight normalized characters and an observation match; successful recovery
can restore the proposed changes. The path also restores truncated target lists when the preserved prefix agrees.
These rules check textual occurrence, not semantic entailment or the completeness of the annotation.

\paragraph{Plan and result handling.}
The query language supports the operations in Equation~\ref{eq:operations}; it does not support arithmetic over values, joins or free-form code.

Compilation uses temperature 0 and no thinking. A catalog provides tool names, the 40 most frequent targets and
the 20 most frequent change properties. Filters are case-insensitive substring matches: lists are disjunctions
and different fields are conjoined. Explicit question bounds define the scope, including open bounds and distinct
bounds for sub-queries. Results are capped at 6,000 characters, with compressed step ranges and grouped outcomes;
failed actions can be separated by error text. A timeline or list query with no matches is retried once with only
its object and action filters, and any result is marked relaxed. Otherwise ungrounded or empty sub-queries are
dropped. If all are dropped, the answer prompt is exactly TM's.

\paragraph{Zero-generation ledger and plan voting.}
\sys-lite keeps the program-derived step, tool and parameter-difference fields, without model annotations.
The self-consistency ablation compiles the greedy plan plus four samples at temperature 0.7 and retains results
supported by at least three identical executed sub-query outputs. These are ablations, not additional components
of the principal \sys{} variant.

\section{Full AMA-Bench Results and Comparison Scope}
\label{app:full}

\begin{figure}[t]
\centering
\includegraphics[width=\columnwidth]{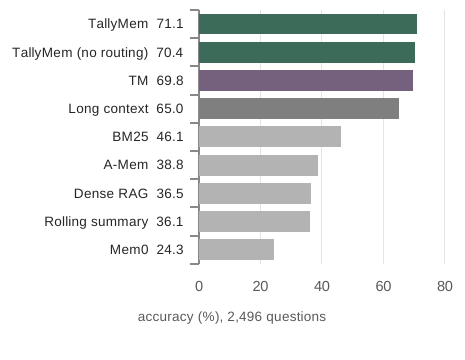}
\caption{Comparison of Qwen3-32B systems on the 2,496 AMA-Bench questions. \sys{} denotes the routed main
system; ``no routing'' is its ablation. External memory systems use the same deployment, with failed memory
constructions counted as wrong. Deployment and adaptation details below define the comparison scope.}
\label{fig:full_systems}
\end{figure}

Table~\ref{tab:full_detail} gives full category results and variants. TM includes the fixed counter described
in Section~\ref{sec:tm}. With Qwen3-8B on our local server (Table~\ref{tab:full}), TM exceeds long context
(\FullEightTmVsLongWL, $\FullEightTmVsLongP$) and \sys{} exceeds TM (\FullEightGateWL, $\FullEightGateP$).
Published AMA scores use different settings: their Qwen3 long-context control is 51.8\%, compared with
\FullLong\% here, and their AMA-Agent scores 57.2\%. These scores provide context; matched comparisons are reported below.
BM25, dense RAG, rolling summary, Mem0 and A-Mem run in the same deployment; Qwen3-32B answers and also builds
their memories, and dense retrieval, Mem0 and A-Mem use all-MiniLM-L6-v2 embeddings. BM25 and dense RAG show their five top-scoring
blocks in original order (2,400-character blocks for BM25, passages within the embedding model's input limit for dense
RAG), much less text than TM's 26k-token view. All five are well below TM; the
best, BM25, scores \FullBlBest\% (TM wins \FullTmVsBlBmWL, $\FullTmVsBlBmP$). Failed memory constructions count as
wrong: A-Mem failed on \FullBlAmemFailedTraj{} of 208 trajectories (\FullBlAmemScored\% on the \FullBlAmemScoredN{}
answered questions) and Mem0 lost \FullBlMemzeroFailed{} questions. Three adaptations were needed to run the released
code: rolling summary states its 1,500-word budget after the history block, retries once at 750 words and otherwise
keeps the previous summary (\FullBlRollSkipped{} blocks skipped in \FullBlRollSkippedTraj{} trajectories); Mem0 gets a
separate local store per worker; A-Mem receives JSON without Markdown fences.

\begin{table*}[t]
\centering\small
\begin{tabular}{lcccccc}
\toprule
System & All & A & B & C & D & vs.\ TM: wins/losses ($p$) \\
\midrule
\multicolumn{7}{l}{\emph{Qwen3-32B reader, official judge, local bf16 deployment}} \\
No trajectory (question only) & 21.9 & 21.6 & 28.5 & 20.2 & 15.5 & +115/$-$1311 ($<$0.001) \\
Long context & 65.0 & 65.2 & 72.0 & 63.4 & 57.2 & +262/$-$381 ($<$0.001) \\
TM & 69.8 & 72.2 & 72.8 & 73.1 & 55.3 & -- \\
BM25 (baseline suite) & 46.1 & 47.9 & 54.5 & 42.3 & 36.2 & +172/$-$763 ($<$0.001) \\
Dense RAG (baseline suite) & 36.5 & 39.0 & 46.6 & 30.8 & 25.6 & +165/$-$997 ($<$0.001) \\
Rolling summary (baseline suite) & 36.1 & 36.4 & 40.8 & 32.8 & 34.3 & +152/$-$992 ($<$0.001) \\
Mem0 (baseline suite) & 24.3 & 24.9 & 28.4 & 21.9 & 20.8 & +153/$-$1289 ($<$0.001) \\
A-Mem (baseline suite) & 38.8 & 40.6 & 47.3 & 34.2 & 30.2 & +172/$-$945 ($<$0.001) \\
\sys$^\S$ & 71.1 & 72.2 & 72.8 & 73.4 & 62.8 & +47/$-$14 ($<$0.001) \\
\sys-lite, no routing & 70.8 & 71.8 & 73.0 & 73.4 & 61.4 & +51/$-$27 (0.0088) \\
\sys, no routing & 70.4 & 71.0 & 73.5 & 72.5 & 61.6 & +79/$-$63 (0.21) \\
\sys{} refined compiler$^\dagger$, no routing & 70.5 & 71.5 & 72.3 & 72.2 & 63.0 & +76/$-$59 (0.17) \\
\sys, rule-routed$^\ddagger$ & 70.8 & 71.5 & 73.0 & 73.4 & 62.3 & +52/$-$26 (0.0043) \\

\midrule
\multicolumn{7}{l}{\emph{gpt-oss-120b reader, same judge model, DeepInfra}} \\
Long context & 73.6 & 72.9 & 81.4 & 68.9 & 71.0 & +205/$-$324 ($<$0.001) \\
TM & 78.4 & 80.0 & 83.9 & 75.4 & 71.7 & -- \\
\sys$^\S$ & 79.2 & 80.5 & 84.1 & 75.1 & 76.1 & +40/$-$19 (0.0086) \\
\sys-lite, no routing & 78.7 & 79.6 & 84.2 & 75.1 & 74.6 & +36/$-$27 (0.31) \\
\sys, no routing & 79.2 & 80.2 & 83.9 & 76.0 & 75.4 & +75/$-$54 (0.078) \\

\bottomrule
\end{tabular}
\caption{Full AMA-Bench, 2,496 questions; accuracy (\%). Categories: A recall, B causal inference,
C state updating, D state abstraction. $^\dagger$Refined compiler: two compile rules added after inspecting the principal compiler's errors.
$^\ddagger$Rule routing: post hoc restriction to pre-defined aggregation families.
$^\S$With the router of Appendix~\ref{app:gate}; ``no routing'' rows attach every nonempty computed evidence.
External baselines: same deployment and reader; failed memory constructions count as wrong.}
\label{tab:full_detail}
\end{table*}

Without routing, on \FullRuleN{} rule-selected questions Qwen3 accuracy improves from \FullRuleTm\% to \FullRuleEv\%
($\FullRuleEvP$). Outside those families, it falls slightly from \FullOtherTm\% to \FullOtherEv\%
($\FullOtherEvP$). Within questions with nonempty computed evidence, the low-load stratum scores
\FullLowTm\% versus \FullLowEv\% ($\FullLowEvP$), while the high-load stratum scores \FullHighTm\%
versus \FullHighEv\% ($\FullHighEvP$). The lite variant (also without routing) improves significantly for Qwen3 ($\FullLiteP$), but
not for gpt-oss ($\FullOssLiteP$). A richer ledger is therefore not a uniformly better representation.

\paragraph{Comparison with reported methods.}
\label{app:reported_methods}
Table~\ref{tab:reported_methods} compares TM and \sys{} with externally reported AMA-Bench methods using
Qwen3-32B. The external results use each method's own deployment, prompts and context limits; the lower block
reports this study's matched runs from Table~\ref{tab:full}. TM exceeds every external score listed, and \sys{}
improves further. For the common long-context baseline, the reported score is 51.35\% and the matched run is
\FullLong\%, reflecting the different settings. Table~\ref{tab:full} provides the controlled comparisons.
\begin{table}[t]
\centering\small
\begin{tabular}{lr}
\toprule
Method & Accuracy (\%) \\
\midrule
\multicolumn{2}{l}{\textit{Externally reported results}} \\
AMA-agent & 56.91 \\
Hipporag2 & 44.05 \\
HybridFocus & 62.46 \\
Long context & 51.35 \\
Memorag & 45.70 \\
Memorybank & 34.50 \\
Qwen3-Embedding-4B & 42.04 \\
TheoryLab & 52.16 \\
VeraKV & 64.78 \\

\midrule
\multicolumn{2}{l}{\textit{Results from this study}} \\
Long context & \FullLong \\
TM & \FullTm \\
TM + \sys{} & \FullGate \\
\bottomrule
\end{tabular}
\caption{AMA-Bench method comparison with Qwen3-32B. External scores are reported under their respective
settings; this study's scores come from Table~\ref{tab:full}.}
\label{tab:reported_methods}
\end{table}

\section{Reader Controls and Rule-Selected Results}
\label{app:readers}

\begin{figure*}[t]
\centering
\includegraphics[width=\textwidth]{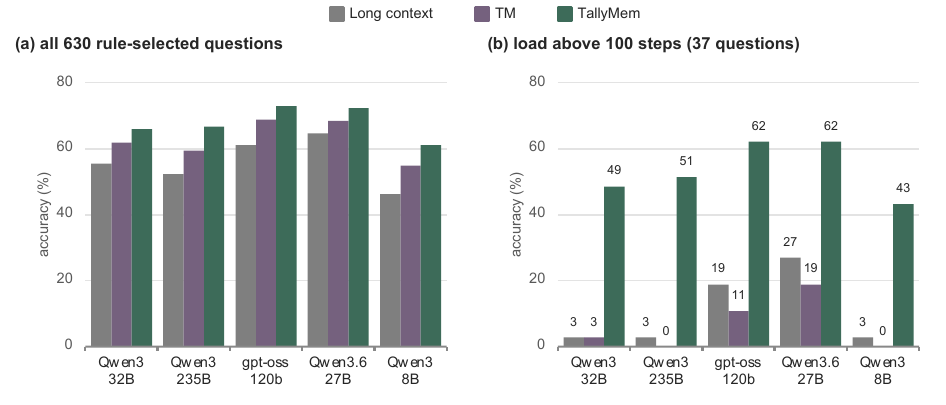}
\caption{Five-reader diagnostic comparison, without routing: (a) all 630 common rule-selected questions;
(b) the 37 questions whose compiled queries match more than 100 steps. Each method's frozen contexts and evidence
are shared across readers, with a common judge. The high-load subset is small; paired tests and full load strata appear in Tables~\ref{tab:readers} and~\ref{tab:load}.}
\label{fig:five_readers}
\end{figure*}

Figure~\ref{fig:five_readers} visualizes the reader comparison and its high-load subset. These are diagnostic
results with no routing, rather than the full-benchmark results of Table~\ref{tab:full}.

\begin{table*}[t]
\centering\small
\begin{tabular}{lcc}
\toprule
Method (Qwen3-32B reader) & correct & vs.\ TM: +/$-$ ($p$) \\
\midrule
No trajectory (question only) & 95/630 & +44/$-$339 ($<$0.001) \\
Long context & 350/630 & +61/$-$101 (0.002) \\
TM & 390/630 & -- \\
TM + enumerate-then-count prompt & 386/630 & +43/$-$47 (0.75) \\
\sys-lite (zero-generation ledger) & 407/630 & +32/$-$15 (0.02) \\
\sys & 416/630 & +55/$-$29 (0.006) \\
\sys{} + plan self-consistency & 414/630 & +50/$-$26 (0.008) \\
\sys{} refined compiler$^\dagger$ & 423/630 & +61/$-$28 ($<$0.001) \\

\bottomrule
\end{tabular}
\caption{Correct answers on the 630 rule-selected questions. $^\dagger$Refined compiler adds two compile rules chosen after
inspecting the principal compiler's errors (Section~\ref{sec:analysis}).}
\label{tab:main}
\end{table*}

\begin{table*}[t]
\centering\small
\begin{tabular}{lcccccc}
\toprule
Answer model & $n$ & Long context & TM & \sys-lite & \sys & \sys{} vs.\ TM \\
\midrule
Qwen3-32B (thinking) & 630 & 55.6 & 61.9 & 64.6 & 66.0 & +55/$-$29 (0.006) \\
Qwen3-235B-A22B-2507 & 630 & 52.4 & 59.5 & 64.3 & 66.8 & +78/$-$32 ($<$0.001) \\
gpt-oss-120b (reasoning) & 630 & 61.1 & 68.9 & 70.2 & 73.0 & +54/$-$28 (0.005) \\
Qwen3.6-27B (thinking)$^\S$ & 630 & 64.6 & 68.4 & 72.7 & 72.4 & +51/$-$26 (0.006) \\
Qwen3-8B (thinking)$^\S$ & 630 & 46.3 & 54.9 & 60.3 & 61.1 & +61/$-$22 ($<$0.001) \\

\bottomrule
\end{tabular}
\caption{Five answer models on the same frozen contexts and evidence (accuracy \%, paired sign tests; the judge model
is Qwen3-32B throughout). $^\S$Answered on our local server (bf16); the others on DeepInfra.}
\label{tab:readers}
\end{table*}

\begin{table}[t]
\centering\small
\begin{tabular}{lc}
\toprule
Family & TM $\to$ \sys{} ($p$) \\
\midrule
Count / frequency (F1) & 152 $\to$ 176 / 246 ($<$0.001) \\
First / last change (F2) & 91 $\to$ 92 / 152 (1.00) \\
State tracing (F3) & 113 $\to$ 115 / 186 (0.83) \\
Action enumeration (F4) & 52 $\to$ 72 / 91 ($<$0.001) \\

\bottomrule
\end{tabular}
\caption{TM $\to$ \sys{} correct / questions by rule-defined family on the 630 rule-selected questions ($p$ from a
paired sign test; one question may be in several families).}
\label{tab:families}
\end{table}

\begin{table*}[t]
\centering\small
\setlength{\tabcolsep}{4pt}
\begin{tabular}{lcc}
\toprule
Answer model & trajectory fits & does not fit \\
\midrule
Qwen3-32B (thinking) & 66.0 / 69.6 / 75.1 (0.01) & 45.5 / 54.5 / 57.3 (0.21) \\
Qwen3-235B-A22B-2507 & 62.5 / 68.0 / 73.5 (0.02) & 42.7 / 51.4 / 60.4 ($<$0.001) \\
gpt-oss-120b (reasoning) & 75.4 / 80.9 / 82.2 (0.64) & 47.4 / 57.3 / 64.2 ($<$0.001) \\
Qwen3.6-27B (thinking)$^\S$ & 79.6 / 77.3 / 79.9 (0.24) & 50.2 / 59.8 / 65.1 (0.01) \\
Qwen3-8B (thinking)$^\S$ & 56.6 / 58.6 / 65.4 (0.001) & 36.4 / 51.4 / 57.0 (0.010) \\

\bottomrule
\end{tabular}
\caption{Long context / TM / \sys{} accuracy (\%) by nominal raw-history fit (630 questions;
$p$: \sys{} vs.\ TM). When it does not fit, long context collapses (retrieval bottleneck);
when it fits, computation adds significantly for three of the five readers. Fit does not guarantee full TM field
visibility (Appendix~\ref{app:visibility}). $^\S$Local server.}
\label{tab:fit}
\end{table*}

\subsection{Aggregation Load and Prompting Controls}
\begin{table*}[t]
\centering\small
\begin{tabular}{lccc}
\toprule
Answer model & load 1--15 & load 16--100 & load $>$100 \\
\midrule
Qwen3-32B (thinking) & 63.2 / 72.9 / 71.2 & 33.3 / 44.1 / 57.8 & 2.7 / 2.7 / 48.6 \\
Qwen3-235B-A22B-2507 & 58.9 / 68.6 / 74.2 & 33.3 / 48.0 / 57.8 & 2.7 / 0.0 / 51.4 \\
gpt-oss-120b (reasoning) & 61.5 / 75.3 / 76.9 & 57.8 / 64.7 / 66.7 & 18.9 / 10.8 / 62.2 \\
Qwen3.6-27B (thinking)$^\S$ & 67.6 / 75.9 / 77.6 & 60.8 / 57.8 / 61.8 & 27.0 / 18.9 / 62.2 \\
Qwen3-8B (thinking)$^\S$ & 53.2 / 69.6 / 72.2 & 24.5 / 25.5 / 40.2 & 2.7 / 0.0 / 43.2 \\

\bottomrule
\end{tabular}
\caption{Accuracy (\%) as long context / TM / \sys{} by aggregation load. Of the 630 diagnostic questions, 438 have defined load; counts are \ReaderLoadSizes\unskip{} for the three strata. $^\S$Local server.}
\label{tab:load}
\end{table*}
\paragraph{Prompting helps, but not at high load.}
We add one sentence to TM's answer instructions: list every relevant step with its step number, then compute the
answer from that list. For Qwen3-32B, 
on the same 630 questions, accuracy is not significantly different: 61.9\% without and 61.3\% with the instruction (+43/$-$47, $p=0.75$). It stays at 2.7\% above 100 steps (2.7\% without).
\unskip{}
Table~\ref{tab:enumerate} repeats this for the other two readers: for the reasoning model gpt-oss-120b it moves overall accuracy from 68.9\% to 74.0\% (+61/$-$29, $p<0.001$; \sys{} 73.0\%), but above 100 steps it reaches only 18.9\% (\sys{} 62.2\%); for the non-thinking Qwen3-235B it moves overall accuracy from 59.6\% to 61.2\% (+77/$-$67, $p=0.45$; \sys{} 66.9\%), but above 100 steps it reaches only 24.3\% (\sys{} 51.4\%).
\unskip{} The instruction works best for the reasoning reader at small and
moderate loads, where it even exceeds \sys{} (80.6\% vs.\ 76.9\% at 1--15 steps, 71.6\% vs.\ 66.7\% at 16--100).
It does not help the non-thinking reader at those loads. For all three readers, high-load accuracy remains low.

\begin{table*}[t]
\centering\small
\resizebox{\textwidth}{!}{%
\begin{tabular}{lccc}
\toprule
Answer model & all & +enum.\ vs.\ TM & load $>$100 \\
\midrule
Qwen3-32B (thinking) & 61.9 / 61.3 / 66.0 / 65.4 & +43/$-$47 (0.75) & 2.7 / 2.7 / 48.6 / 56.8 \\
Qwen3-235B-A22B-2507$^\ast$ & 59.6 / 61.2 / 66.9 / 65.3 & +77/$-$67 (0.45) & 0.0 / 24.3 / 51.4 / 62.2 \\
gpt-oss-120b (reasoning) & 68.9 / 74.0 / 73.0 / 76.3 & +61/$-$29 ($<$0.001) & 10.8 / 18.9 / 62.2 / 67.6 \\

\bottomrule
\end{tabular}}
\caption{Enumeration controls on 630 diagnostic questions (37 with load above 100). Accuracy columns (\%): TM / TM with enumeration / \sys{} / \sys{} with enumeration. The comparison column gives wins/losses and sign-test $p$ for TM with enumeration versus TM. $^\ast$629 of 630: one question failed four times with a transport error.}
\label{tab:enumerate}
\end{table*}

\paragraph{Execution and enumeration combine.}
Adding the enumeration instruction to \sys's prompt gives gpt-oss-120b 76.3\% (630 questions), against 73.0\% for \sys{} (+60/$-$39, $p=0.04$) and 74.0\% for the instruction alone (+46/$-$31, $p=0.11$), and keeps 67.6\% above 100 steps, where the instruction alone reaches 18.9\%.
\unskip{} For the other two readers the combination does not raise overall
accuracy (65.4\% and 65.3\%, against 66.0\% and 66.9\% for \sys; Table~\ref{tab:enumerate}, last value in each cell),
because the instruction does not help them at small loads. Above 100 steps, however, it helps all three readers:
48.6\% $\to$ 56.8\%, 51.4\% $\to$ 62.2\% and 62.2\% $\to$ 67.6\%.

Larger, reasoning and newer readers and enumeration prompts improve some small and moderate aggregations, but high-load errors remain. Reliable aggregation therefore does not follow automatically from supplying the text or strengthening the reader.

\section{Context Fit and Actual Visibility}
\label{app:visibility}
The context-fit strata use raw trajectory length at most 24k tokens. This is a property of the history, not proof
that TM displays it completely: TM's saved configuration uses per-field caps of 1,200 characters, or 4,000 for
anchor steps. Selecting every step can therefore still omit part of an observation.

For a separate check, we join the existing fit and load annotations to the saved Qwen3-32B long-context answers.
The long-context wrapper renders all original actions and observations and truncates only when the total exceeds
its token target. Every saved context in the nominal-fit, nonempty-load strata is strictly below that target.
Accuracy is \VisibilityLowAcc\% (\VisibilityLowCorrect/\VisibilityLowN) at loads 1--15 and
\VisibilityHighAcc\% (\VisibilityHighCorrect/\VisibilityHighN) above 15.
This establishes a high-load decline with complete visible histories for the long-context control; it does not
establish complete visibility for TM. This is an additional descriptive audit of existing answers, not a new
model evaluation or a pre-registered analysis. The same compiled-load and domain-composition limitations apply.

\section{Robustness of the Aggregation Load}
\label{app:robust}
\paragraph{Zero-generation plans.}
Binning by the steps matched by \sys-lite's independently compiled plan (199 questions with lite evidence):
\begin{itemize}
  \item 1--5 steps: TM 89\%, \sys{} 85\%;
  \item 6--15 steps: 81\% vs.\ 79\%;
  \item above 100 steps: 3\% vs.\ 50\% (+17/$-$1, $p=1.4\times10^{-4}$).
\end{itemize}

\paragraph{Gold-derived loads.}
We parse each gold answer for ``$N$ times''-style counts and cited step numbers, and take the larger. This load
correlates with the compiled load at 0.61. For the 17 questions with a gold load above 100, TM is correct on none and
\sys{} on all. The probe of Appendix~\ref{app:probe} replaces this approximation by event sets parsed from the logs.

\paragraph{Within domains.}
Full benchmark, TM by load bin (question counts in parentheses; bins with fewer than five questions omitted), with
the earlier independent TM run:
\begin{itemize}
  \item embodied: 83\%\,(122), 71\%\,(48), 45\%\,(11), 20\%\,(20), 8\%\,(38); earlier run 81\%, 71\%, 45\%, 30\%, 3\%;
  \item text-to-SQL: 89\%\,(46), 75\%\,(44), 62\%\,(65), 33\%\,(27); earlier run 89\%, 73\%, 66\%, 48\%;
  \item web: 86\%\,(66), 67\%\,(36), 69\%\,(16); earlier run 88\%, 69\%, 56\%.

\end{itemize}

\paragraph{Repeated TM trend.}
A separate TM run gives \FullBinHistA\%, \FullBinHistB\%, \FullBinHistC\%, \FullBinHistD\% and
\FullBinHistE\% over the same load bins. The main figure presents one representative matched comparison;
this independent run is retained as a robustness check rather than a second main-text series.

\section{Controlled Aggregation Probe}
\label{app:probe}

\paragraph{Design.}
A deterministic script generates the questions over all \ProbeEmbEp{} embodied and \ProbeSwEp{} software trajectories of
AMA-Bench, the two domains whose actions have a fixed format. Answers and event sets $S^{\star}$ are parsed from the raw
action and observation strings only, without the ledger, the compiler or a model: embodied actions by their leading
verb; software actions by the tool name before the colon, with editor commands and paths read from the JSON arguments
and the exit code of a bash command taken from the last ``exit code $N$'' in its observation. For a trajectory with $n$
steps the bounds are $b_1=\lfloor n/4\rfloor-1$, $b_2=\lfloor n/2\rfloor-1$ and $b_3=n-1$, phrased ``up to and
including step $b$''. A template is skipped for a trajectory without the relevant event or, for the most-visited
location, with a tie. Each question states its answer format (an integer, a step number, a comma-separated list or
``none'', or a name). Table~\ref{tab:probe} lists the templates and the accuracy of each system.

\paragraph{Protocol.}
Templates, bounds, systems, scoring and analyses were written down before the questions were generated, and the
question files were hashed before any answer was produced. The systems are TM, unchanged; \sys{} in its frozen principal
configuration without routing, compiling on the same deployment; a fixed counter, which runs the same executor and
renderer with one plan per question (group by tool within the stated bound, or over all steps for first-event
templates) and thus involves no compilation; and long-context reading. Three arms were added after these results: a
correct-plan arm, a second reader and free-form code (below). Qwen3-32B (bf16, local vLLM) answers with
thinking enabled; compilation and judging run without thinking, as in the main evaluation.

\paragraph{Scoring.}
The primary score is the official AMA-Bench judge, with the parsed answer as reference. A deterministic extraction was
registered as a secondary score; its integer rule was changed once, to take the leading integer first, after some
answers had been read, and we report it as a post-hoc change. We read every answer on which the judge and the
extraction disagree (\ProbeJudgeReviewed{} of \ProbeJudged): in \ProbeJudgeLenient{} the judge accepted a wrong answer,
typically a count one or two short or, for long context, a statement that a file was never modified or created, and in
the others the extraction erred. With these readings the accuracies are
\ProbeAdjTally\% for \sys, \ProbeAdjTm\% for TM, \ProbeAdjCounter\% for the fixed counter and \ProbeAdjLong\% for
long context, against \ProbeTally\%, \ProbeTm\%, \ProbeCounter\% and \ProbeLong\% under the judge; the comparisons below do not change.
Questions of one trajectory are correlated (E1 and S1 repeat a count at three bounds), so intervals resample whole
trajectories (\ProbeCIClusters{} clusters, 10,000 resamples): under the judge, \sys{} exceeds TM by \ProbeCITallyTm{} points
and the fixed counter by \ProbeCITallyCounter{}, TM exceeds long context by \ProbeCITmLong, and the correct-plan arm exceeds
\sys{} by \ProbeCIOracleTally.

\begin{table*}[t]
\centering\footnotesize
\resizebox{\textwidth}{!}{%
\begin{tabular}{lllrrrrrrrr}
\toprule
& Question & Operation & $n$ & $\overline{|S^{\star}|}$ & \sys & TM & Counter & Long context & Correct plan & Code \\
\midrule
E1 & count \texttt{go to} (three bounds; ``how many times'') & count & 90 & 36 & 100.0 & 100.0 & 100.0 & 78.9 & 100.0 & 95.6 \\
E2 & as E1 at the full bound (``how often'') & count & 30 & 62 & 100.0 & 46.7 & 100.0 & 43.3 & 100.0 & 36.7 \\
E3 & count \texttt{open} & count & 30 & 5 & 100.0 & 96.7 & 100.0 & 73.3 & 100.0 & 100.0 \\
E4 & opens and closes of the most-used receptacle & count & 25 & 4 & 100.0 & 100.0 & 100.0 & 88.0 & 100.0 & 100.0 \\
E5 & distinct objects taken & distinct & 30 & 2 & 100.0 & 100.0 & 100.0 & 86.7 & 100.0 & 93.3 \\
E6 & last step with a \texttt{take} & last & 30 & 2 & 100.0 & 90.0 & 100.0 & 93.3 & 100.0 & 100.0 \\
E7 & list the \texttt{open} steps up to the half bound & list & 30 & 3 & 96.7 & 100.0 & 100.0 & 83.3 & 100.0 & 100.0 \\
E8 & most-visited location & argmax & 30 & 62 & 100.0 & 96.7 & 96.7 & 100.0 & 100.0 & 96.7 \\
E9 & first \texttt{examine} & first & 26 & 3 & 100.0 & 92.3 & 100.0 & 96.2 & 100.0 & 100.0 \\
\addlinespace
S1 & count \texttt{execute\_bash} (three bounds; ``how many times'') & count & 108 & 32 & 99.1 & 98.1 & 94.4 & 3.7 & 100.0 & 92.6 \\
S2 & as S1 at the full bound (``how often'') & count & 36 & 57 & 97.2 & 8.3 & 83.3 & 0.0 & 94.4 & 83.3 \\
S3 & bash commands with a non-zero exit code & count & 36 & 11 & 30.6 & 5.6 & 33.3 & 2.8 & 22.2 & 19.4 \\
S4 & modifications of the most-edited file & count & 36 & 9 & 75.0 & 86.1 & 72.2 & 27.8 & 97.2 & 69.4 \\
S5 & last modification of that file & last & 36 & 9 & 97.2 & 94.4 & 83.3 & 41.7 & 100.0 & 88.9 \\
S6 & distinct paths viewed & distinct & 36 & 21 & 94.4 & 69.4 & 69.4 & 22.2 & 94.4 & 86.1 \\
S7 & first file creation & first & 36 & 7 & 100.0 & 97.2 & 97.2 & 50.0 & 100.0 & 100.0 \\

\bottomrule
\end{tabular}}
\caption{Probe templates and accuracy (\%, official judge; Qwen3-32B). E: embodied, S: software. $n$: questions;
$\overline{|S^{\star}|}$: mean number of logged events in the answer's event set. Counter: the fixed group-by-tool
plan run by \sys's executor. Over all \ProbeN{} questions: \sys{} \ProbeTally, TM \ProbeTm{} (\sys{}
\ProbeTallyVsTmWL, $p$ \ProbeTallyVsTmP), counter \ProbeCounter{} (\ProbeTallyVsCounterWL, $p$ \ProbeTallyVsCounterP).}
\label{tab:probe}
\end{table*}

\paragraph{Bounds and wording.}
E1 and S1 ask the same count at three bounds of one trajectory, so the context is the same and only the load changes;
E2 and S2 repeat the count at the full bound with ``how often'' instead of ``how many times''
(Table~\ref{tab:probe_bounds}). TM's counter fires on ``how many times'', and none of the execution systems then declines
with the bound; long context, which has no counter, does (see \emph{Long context} below). With ``how often'' TM falls to \ProbeWordEmbTmPlain\% and \ProbeWordSwTmPlain\%, whereas \sys{} and the
fixed counter keep their accuracy.

\paragraph{Long context.}
Long-context reading uses the official format and truncation, which keeps the first 70\% and the last 30\% of the token
budget. The whole history fits for \ProbeLongFitsEmb{} embodied questions and for no software question; every step of
$S^{\star}$ is visible for \ProbeLongCompleteN{} questions (\ProbeLongCompleteEmb{} embodied, \ProbeLongCompleteSw{}
software). Its failures on software trajectories are therefore mostly failures of access; where every counted step is
visible it still miscounts, mostly by counting too many. On embodied questions with complete access its accuracy is
\ProbeLongVisEmbBinA\%, \ProbeLongVisEmbBinB\%, \ProbeLongVisEmbBinC\% and \ProbeLongVisEmbBinD\% over the four load
bins of Figure~\ref{fig:load} (\ProbeLongVisEmbBinNA, \ProbeLongVisEmbBinNB, \ProbeLongVisEmbBinNC{} and
\ProbeLongVisEmbBinND{} questions). On the \ProbeLongVisEp{} trajectories whose counted steps are visible at all three
bounds, E1 falls from \ProbeLongVisBOne\% to \ProbeLongVisBTwo\% and \ProbeLongVisBThree\% as the mean load grows from
\ProbeLongVisLoadOne{} to \ProbeLongVisLoadTwo{} and \ProbeLongVisLoadThree{} events.

\begin{table}[t]
\centering\footnotesize
\resizebox{\columnwidth}{!}{%
\begin{tabular}{llrrrrr}
\toprule
& Bound and wording & $\overline{|S^{\star}|}$ & \sys & TM & Counter & Long \\
\midrule
E1 & $b_1$, ``how many times'' & 16 & 100.0 & 100.0 & 100.0 & 100.0 \\
E1 & $b_2$, ``how many times'' & 31 & 100.0 & 100.0 & 100.0 & 80.0 \\
E1 & $b_3$, ``how many times'' & 62 & 100.0 & 100.0 & 100.0 & 56.7 \\
E2 & $b_3$, ``how often'' & 62 & 100.0 & 46.7 & 100.0 & 43.3 \\
\addlinespace
S1 & $b_1$, ``how many times'' & 13 & 100.0 & 100.0 & 97.2 & 5.6 \\
S1 & $b_2$, ``how many times'' & 26 & 100.0 & 97.2 & 97.2 & 2.8 \\
S1 & $b_3$, ``how many times'' & 57 & 97.2 & 97.2 & 88.9 & 2.8 \\
S2 & $b_3$, ``how often'' & 57 & 97.2 & 8.3 & 83.3 & 0.0 \\

\bottomrule
\end{tabular}}
\caption{Same count, three bounds and two wordings (accuracy, \%; official judge).}
\label{tab:probe_bounds}
\end{table}

\paragraph{Correct plans.}
To separate specification errors from the rest, a supplementary arm replaces each compiled plan by a fixed plan written
from the template's definition, with parameters taken only from the question (the bound, the receptacle or the file),
and keeps the executor, renderer, reader and judge. The arm was designed after inspecting \sys's misses on S3--S5, so
it is diagnostic rather than confirmatory. Its accuracy is \ProbeOracle\% (\ProbeOracleVsTallyWL{} against \sys, $p$
\ProbeOracleVsTallyP), and \ProbeOracleNoSThree\% without S3. On S4 it reaches \ProbeOracleSFour\% against
\ProbeSFourTally\% (\ProbeOracleSFourWL, $p$ \ProbeOracleSFourP). S3 stays at \ProbeOracleSThree\%: no plan can express
a non-zero exit code over this ledger, and the closest one, the failure outcome, selects the wrong events
(Table~\ref{tab:probe_event}, last column: questions whose correct plan selects exactly $S^{\star}$).

\paragraph{Second reader.}
With Qwen3-8B answering from the same evidence (and Qwen3-32B judging), \sys{} reaches \ProbeEightTally\%, TM
\ProbeEightTm\%, the fixed counter \ProbeEightCounter\%, long context \ProbeEightLong\% and the correct-plan arm
\ProbeEightOracle\% (Table~\ref{tab:probe_readers}). \sys{} exceeds TM by \ProbeEightCITallyTm{} points, and TM's
dependence on the wording persists (\ProbeEightWordEmbTmTrig\% against \ProbeEightWordEmbTmPlain\% on embodied counts,
\ProbeEightWordSwTmTrig\% against \ProbeEightWordSwTmPlain\% on software counts). Two results do not carry over. With
this reader \sys{} and the fixed counter are level (\ProbeEightTallyVsCounterWL{} for \sys). On the \ProbeLongVisEp{}
fully visible trajectories, long context answers E1 correctly for \ProbeEightLongVisBOne\%, \ProbeEightLongVisBTwo\%
and \ProbeEightLongVisBThree\% of them at the three bounds, a weaker decline than with Qwen3-32B, while \sys{} falls to
\ProbeEightTallyVisBThree\% at the highest bound: in each of these errors the reader recounts the logged steps instead
of using the computed count. Computed evidence helps a weaker reader only when the reader adopts it. The judge
accepted \ProbeEightJudgeLenient{} wrong answers among the \ProbeEightJudgeReviewed{} we read for this reader.

\begin{table}[t]
\centering\footnotesize
\resizebox{\columnwidth}{!}{%
\begin{tabular}{lrrrrr}
\toprule
Reader & \sys & TM & Counter & Long context & Correct plan \\
\midrule
Qwen3-32B & 93.8 & 82.8 & 89.9 & 49.3 & 94.9 \\
Qwen3-8B & 83.1 & 70.1 & 81.9 & 54.3 & 84.3 \\

\bottomrule
\end{tabular}}
\caption{Probe accuracy (\%, official judge) by reader; all arms use the same evidence.}
\label{tab:probe_readers}
\end{table}

\paragraph{Free-form code.}
A final arm replaces the compiled plan by a program, using the prompt, sandbox and single repair of
Appendix~\ref{app:code} unchanged; the program sees the ledger rows and every step verbatim, observations included, and
is generated by the same model without thinking. Programs give evidence for \ProbeCodeRendered{} of the \ProbeN{}
questions. With the Qwen3-32B reader the arm reaches \ProbeCode\% (\sys{} \ProbeTally\%, \ProbeCodeVsTallyWL{} for code,
$p$ \ProbeCodeVsTallyP; interval for code minus \sys: \ProbeCICodeTally{} points), \ProbeCodeEmb\% on embodied and
\ProbeCodeSw\% on software trajectories. On non-zero exit codes (S3), which code could parse from the observations, it
reaches \ProbeCodeSThree\% against \ProbeSThreeTally\% for \sys: \ProbeCodeSThreeObs{} of its \ProbeSThreeN{} programs read the
observations, \ProbeCodeSThreeQuote{} parse the ledger's short quotes and \ProbeCodeSThreeOutcome{} count failure
outcomes; on file modifications (S4) \ProbeCodeSFour\% against
\ProbeSFourTally\%.

\paragraph{Event sets.}
Table~\ref{tab:probe_event} compares \sys's evidence with $S^{\star}$. A question counts as exact when some sub-query or
group of its evidence selects exactly $S^{\star}$; precision and recall are those of the candidate set with the highest
F1 (zero without evidence), which favours the system. Over the \ProbeEventNonEmpty{} questions with a nonempty
$S^{\star}$, \ProbeEventExact{} are exact (F1 \ProbeEventFOne). The misses concentrate in S3--S5. S3 asks for non-zero
exit codes, which the ledger does not record; its plans use the failure outcome instead, which labels some runs that
exit with code 0 as failures and some non-zero exits, such as \texttt{grep} without a match, as successes. In S4 and S5
every inexact plan adds \texttt{view} to the modifying commands that the question names; readers still answer
\ProbeSFourWrongSetCorrect{} of the \ProbeSFourWrongSet{} affected S4 questions correctly from the listed steps. For the
fixed counter, the per-tool outcome counts it renders are not candidate sets, so its S3 row understates its evidence.

\begin{table}[t]
\centering\footnotesize
\resizebox{\columnwidth}{!}{%
\begin{tabular}{lrrrrrrr}
\toprule
& $n$ & Exact & P & R & F1 & Counter F1 & Expressible \\
\midrule
E1 & 90 & 90 & 100.0 & 100.0 & 100.0 & 100.0 & 90/90 \\
E2 & 30 & 30 & 100.0 & 100.0 & 100.0 & 100.0 & 30/30 \\
E3 & 25 & 25 & 100.0 & 100.0 & 100.0 & 100.0 & 30/30 \\
E4 & 25 & 21 & 85.3 & 88.0 & 86.0 & 50.3 & 24/25 \\
E5 & 30 & 30 & 100.0 & 100.0 & 100.0 & 100.0 & 30/30 \\
E6 & 30 & 30 & 100.0 & 100.0 & 100.0 & 100.0 & 30/30 \\
E7 & 24 & 24 & 100.0 & 100.0 & 100.0 & 100.0 & 30/30 \\
E8 & 30 & 29 & 96.7 & 96.7 & 96.7 & 100.0 & 30/30 \\
E9 & 26 & 26 & 100.0 & 100.0 & 100.0 & 100.0 & 26/26 \\
S1 & 108 & 105 & 99.3 & 100.0 & 99.6 & 99.8 & 105/108 \\
S2 & 36 & 35 & 99.6 & 100.0 & 99.8 & 99.8 & 35/36 \\
S3 & 36 & 11 & 82.9 & 91.3 & 85.0 & 32.7 & 11/36 \\
S4 & 36 & 15 & 71.2 & 100.0 & 79.9 & 68.4 & 36/36 \\
S5 & 36 & 15 & 71.2 & 100.0 & 79.9 & 68.4 & 36/36 \\
S6 & 36 & 36 & 100.0 & 100.0 & 100.0 & 100.0 & 36/36 \\
S7 & 36 & 36 & 100.0 & 100.0 & 100.0 & 100.0 & 36/36 \\

\bottomrule
\end{tabular}}
\caption{Event sets of \sys's evidence against $S^{\star}$ (questions with a nonempty $S^{\star}$; P, R and F1 in \%).}
\label{tab:probe_event}
\end{table}

\section{Scope and Robustness of the Gains}
\label{app:gains}

\subsection{Full-Set Gains and Trajectory Splits}
Across five readers, \sys{} improves TM by 4.0--7.3 points on the 630 rule-selected diagnostic questions (Appendix~\ref{app:readers}). Table~\ref{tab:gains} reports gains on the full benchmark and diagnostic set, with trajectory splits for the full benchmark; intervals resample whole trajectories (\RobClusters{} trajectories, \RobBoot{} resamples). The controlled probe and MINTEval provide complementary tests of event aggregation.

\begin{table*}[t]
\centering\footnotesize
\resizebox{\textwidth}{!}{%
\begin{tabular}{lccc}
\toprule
Reader & All & Development half & Held-out half \\
\midrule
Qwen3-32B & $+$1.3 [$+$0.7, $+$2.0] & $+$1.4 [$+$0.6, $+$2.2] & $+$1.3 [$+$0.4, $+$2.2] \\
gpt-oss-120b & $+$0.8 [$+$0.2, $+$1.4] & $+$0.6 [$-$0.2, $+$1.4] & $+$1.1 [$+$0.3, $+$2.0] \\
Qwen3.6-27B & $+$0.7 [$+$0.1, $+$1.4] & $+$1.4 [$+$0.5, $+$2.3] & $+$0.1 [$-$0.7, $+$1.0] \\
Qwen3-8B & $+$1.5 [$+$0.9, $+$2.1] & $+$1.8 [$+$1.0, $+$2.7] & $+$1.2 [$+$0.4, $+$2.1] \\
\addlinespace
Qwen3-32B (630) & $+$4.1 [$+$1.5, $+$6.7] &  &  \\
Qwen3-235B (630) & $+$7.3 [$+$4.0, $+$10.6] &  &  \\
gpt-oss-120b (630) & $+$4.1 [$+$1.1, $+$7.1] &  &  \\
Qwen3.6-27B (630) & $+$4.0 [$+$1.2, $+$6.8] &  &  \\
Qwen3-8B (630) & $+$6.2 [$+$3.6, $+$8.8] &  &  \\

\bottomrule
\end{tabular}}
\caption{Gain of \sys{} over TM in accuracy points, with 95\% intervals from a bootstrap over trajectories. Upper
block: the full benchmark (2,496 questions), routed \sys. Lower block: the 630 diagnostic questions, \sys{} without
routing. Halves follow the trajectory split of Appendix~\ref{app:protocol}; method development used the first.}
\label{tab:gains}
\end{table*}

\subsection{Robustness Checks}

\paragraph{Load-threshold comparison.}
Table~\ref{tab:threshold} reports full-benchmark accuracy across load thresholds.

\begin{table}[t]
\centering\footnotesize
\setlength{\tabcolsep}{3pt}
\resizebox{\columnwidth}{!}{%
\begin{tabular}{lrrrrrrrr}
\toprule
Reader & none & $>$5 & $>$10 & $>$15 & $>$20 & $>$40 & $>$100 & TM \\
\midrule
Qwen3-32B & 70.4 & 70.8 & 71.0 & 71.1 & 71.0 & 71.0 & 70.6 & 69.8 \\
gpt-oss-120b & 79.2 & 79.2 & 79.3 & 79.2 & 79.2 & 78.9 & 79.1 & 78.4 \\
Qwen3.6-27B & 76.8 & 76.7 & 76.7 & 76.6 & 76.6 & 76.7 & 76.5 & 75.9 \\
Qwen3-8B & 62.4 & 61.9 & 62.2 & 62.3 & 62.1 & 61.9 & 61.4 & 60.7 \\

\bottomrule
\end{tabular}}
\caption{Full-benchmark accuracy (\%) when computed evidence is attached only above a load threshold (``none'': always;
TM: never).}
\label{tab:threshold}
\end{table}

\paragraph{Fixed counters and file questions.}
On the probe, the fixed group-by-tool counter matches \sys{} on embodied trajectories (\ProbeEmbCounter\% and
\ProbeEmbTally\%), whose questions name the logged action verbs; \sys{} is ahead on software trajectories
(\ProbeSwTally\% against \ProbeSwCounter\%). On counts of modifications to one file (S4), TM, whose file-event index
groups edits by path, answers \ProbeSFourTm\% against \ProbeSFourTally\% for \sys. On non-zero exit codes (S3), both
execution systems fail on most questions, and so does the correct-plan arm, because the ledger does not record exit
codes (Table~\ref{tab:probe}). On AMA-Bench, TM's own keyword-triggered counter fires on \NoCtrTrig{} of the 2,496
questions. Removing only its computed counts from TM's context, with the evidence unchanged, changes TM on these
questions by \NoCtrSessCI{} points in a same-session rerun (\NoCtrSessWL, $\NoCtrSessP$); the rerun agrees with the
original run on \NoCtrAgree\% of judgments. With the original answers on the other \NoCtrUntrig{} questions, whose
prompts do not change, TM without the counter scores \NoCtrTm\% (\NoCtrVsTmCI{} points against TM, \NoCtrVsTmWL,
$\NoCtrVsTmP$) and exceeds long context by \NoCtrVsLongCI{} points; where the counter never fires, TM exceeds long
context by \NoCtrUntrigCI{} points.

\paragraph{Execution.}
Re-executing the stored plans of all \RobReexecIdentical{} full-benchmark questions whose plan parsed reproduces their
evidence byte for byte. An independently written matcher, which implements the documented filter semantics
(case-insensitive substring terms, outcome sets, inclusive step ranges and the relaxation rule), recomputes the matched
steps, first and last steps and tool or outcome groups of all \RobReexecQueries{} rendered sub-queries of the
\RobReexecRecords{} questions with evidence identically. The evidence is therefore a deterministic function of the plan
and the ledger; whether either is right is what Table~\ref{tab:errors} labels.

\paragraph{Judging, references and development.}
AMA model judging and reference-answer ambiguities add uncertainty. On the probe, where references are parsed from the
logs, the official judge accepted \ProbeJudgeLenient{} wrong answers among the answers we read (Appendix~\ref{app:probe}).
The reference errors found in the error analysis were not used to re-score any result. The later compiler refinement
followed inspection of the principal method's errors, and the reported AMA results include the trajectories used for
method development (Table~\ref{tab:gains} separates the two halves).

\paragraph{Other limitations.}
Aggregation load on AMA-Bench is defined by compiled queries and the extreme bins are small; alternative measures are
checks, not independent difficulty annotations, and the probe's parsed load covers two domains and generated templates.
Quote checks do not establish event semantics, and the executor can use changes lacking a matching-value flag. External
memory baselines were run with one reader and the adaptations of Appendix~\ref{app:full}. The free-form code control
uses one model and one prompt design (Appendix~\ref{app:code}). The error analysis covers one reader on the diagnostic questions,
and its labels were assigned by a single annotator following a fixed procedure, with a written rationale for each case.
TM is below long context on MINTEval, where \sys{} leaves multi-hop accuracy unchanged and long context remains
slightly better on counting questions.

\section{Router Definition}
\label{app:gate}
The router determines $\gamma$ in Equation~\ref{eq:augmentation} from aggregation load and matched-step coverage.
Let $V_q$ contain the step IDs shown without truncation in TM's rendered context. Coverage is
\begin{equation}
 c(q,H)=\mathbf{1}\!\left\{\bigcup_{j\in J}S_j\subseteq V_q\right\}.
 \label{eq:coverage}
\end{equation}
This signal uses step and truncation markers. The router uses
\begin{equation}
\begin{aligned}
 \gamma_{\mathrm{load}}(q,H)&=\mathbf{1}\{\ell(q,H)>15\},\\
 \gamma_{\mathrm{cov}}(q,H)&=\mathbf{1}\{\ell(q,H)>15\ \lor\ c(q,H)=0\}.
\end{aligned}
\label{eq:gates}
\end{equation}
\sys{} uses $\gamma_{\mathrm{load}}$ on AMA-Bench and $\gamma_{\mathrm{cov}}$ on MINTEval; diagnostics set $\gamma=1$.
Load is measured before evidence rendering. The router adds no model call.
The MINTEval rule was fixed on development repositories.
Table~\ref{tab:gate} compares the full-benchmark configurations.

\begin{table*}[t]
\centering\small
\begin{tabular}{llccc}
\toprule
Reader & Questions & TM & no routing & \sys \\
\midrule
Qwen3-32B (server) & all 2,496 & 69.8 & 70.4 (+79/$-$63) & 71.1 (+47/$-$14) \\
 & rule-selected 670 & 64.9 & 68.8 (+52/$-$26) & 69.7 (+39/$-$7) \\
 & other 1,826 & 71.6 & 71.0 (+27/$-$37) & 71.6 (+8/$-$7) \\
\addlinespace
gpt-oss-120b (DeepInfra, pre-registered) & all 2,496 & 78.4 & 79.2 (+75/$-$54) & 79.2 (+40/$-$19) \\
 & rule-selected 670 & 70.2 & 74.0 (+54/$-$28) & 73.3 (+33/$-$12) \\
 & other 1,826 & 81.4 & 81.1 (+21/$-$26) & 81.4 (+7/$-$7) \\
\addlinespace
Qwen3.6-27B (server) & all 2,496 & 75.9 & 76.8 (+74/$-$50) & 76.6 (+38/$-$20) \\
 & rule-selected 670 & 69.6 & 73.3 (+51/$-$26) & 72.5 (+34/$-$14) \\
 & other 1,826 & 78.2 & 78.1 (+23/$-$24) & 78.1 (+4/$-$6) \\
\addlinespace
Qwen3-8B (server, pre-registered) & all 2,496 & 60.7 & 62.4 (+101/$-$59) & 62.3 (+50/$-$12) \\
 & rule-selected 670 & 55.2 & 61.0 (+61/$-$22) & 59.9 (+37/$-$6) \\
 & other 1,826 & 62.8 & 62.9 (+40/$-$37) & 63.1 (+13/$-$6) \\

\bottomrule
\end{tabular}
\caption{Accuracy (\%) and wins/losses against TM with and without the router on full AMA-Bench.}
\label{tab:gate}
\end{table*}

\section{Component Ablations and Cost}
\label{app:ablation}
The upper block in Table~\ref{tab:ablation} uses 630 common questions. The lower block contains earlier development
controls with their own matched TM contexts, including a 14k-token context for the full-ledger test. These groups
must not be treated as a single experiment. Removing quote recovery or relaxed retry changes at most one answer;
sampling five plans does not improve the principal system. Lite keeps much of the gain without annotated fields,
while richer annotations enable queries whose targets or changes are absent from the lite representation.

\begin{table*}[t]
\centering\small
\setlength{\tabcolsep}{4pt}
\begin{tabular}{lccc}
\toprule
Variant & correct & vs.\ \sys & vs.\ TM \\
\midrule
\sys{} & 416/630 & -- & +55/$-$29 (0.006) \\
no quote recovery & 415/630 & +2/$-$3 (1.00) & +55/$-$30 (0.009) \\
no relaxed retry & 415/630 & +2/$-$3 (1.00) & +52/$-$27 (0.007) \\
zero-generation ledger (\sys-lite) & 407/630 & +28/$-$37 (0.32) & +32/$-$15 (0.02) \\
plan self-consistency ($k{=}5$) & 414/630 & +28/$-$30 (0.90) & +50/$-$26 (0.008) \\

\midrule
\multicolumn{4}{l}{\emph{earlier development controls, 217 questions}} \\
TM 14k + full ledger in context & 129 vs.\ 135 & & +16/$-$22 (0.42) \\
single-operation compiler & 136 vs.\ 133 & & +17/$-$14 (0.72) \\
\bottomrule
\end{tabular}
\caption{Ablations (Qwen3-32B). Upper block: 630 rule-selected questions. Lower block: earlier development variants
compared with their own TM context.}
\label{tab:ablation}
\end{table*}

\paragraph{Measured system cost.}
TM and \sys-lite require no construction model calls. The annotated ledger averages 5.5 calls and 42k tokens per
trajectory. Online compilation averages about 1.3k tokens for \sys{} and 0.9k for lite. Computed evidence adds
288 prompt tokens on average in the diagnostic comparison; reader output decreases from 1,302 to 1,007 tokens.
The voting ablation uses five compilation calls. In the matched suite, memory construction takes a median of
\FullBlMemzeroCalls{} model calls per trajectory for Mem0, \FullBlAmemCalls{} for A-Mem and \FullBlRollCalls{} for rolling summary.

\section{Free-Form Code Control}
\label{app:code}
The control replaces only the specification language: instead of a plan in the query language of
Section~\ref{sec:compile}, the model writes a Python program. The prompt keeps \sys's catalog (tools, the 40 most
frequent targets and the 20 most frequent change properties), applicability rule and step-bound rule, and describes
two read-only inputs: the ledger rows (tool, normalized action type, raw action, targets, outcome, quote, changes and
parameter difference) and the verbatim steps with their observations. Qwen3-32B writes the program with thinking off
at temperature 0, as for compilation. Programs run in an operating-system sandbox without network access, file writes
or access to user files, with a whitelist of pure standard-library modules, 10\,s of CPU time and 200k characters of
output. A rejected, failing, timed-out or truncated program is returned once with its error for repair. The printed
output, within the same 6,000-character budget and together with the program text, is added to TM's context as in
Equation~\ref{eq:augmentation} with $\gamma=1$; a program without output leaves TM's answer. The answer model, prompt
and judge are those of the diagnostic runs on the \CodeN{} rule-selected questions. Two secondary arms, specified
before any answers were scored and reported regardless of outcome, give the program the ledger rows only or let the
model think before writing it.

\paragraph{Results.}
The primary arm produces evidence for \CodeEvidence{} questions (\sys: \CodeTallyEvidence); \CodeRepaired{} programs ran after repair and \CodeFailed{} still failed.
Table~\ref{tab:code} compares the three code variants with TM and \sys{}. Across the three code variants, overall accuracy is 60.2--62.1\%, compared with 61.9\% for TM and 66.0\% for \sys{}. At load above 100, code reaches 32.4--45.9\%, below \sys{} at 48.6\%.
\begin{table*}[t]
\centering\small
\setlength{\tabcolsep}{5pt}
\begin{tabular}{lcc}
\toprule
 & All & Load $>$100 \\
\multicolumn{1}{l}{Questions} & 630 & 37 \\
\midrule
TM & 61.9 & 2.7 \\
\sys{} & 66.0 & 48.6 \\
\midrule
Free-form code (ledger + log) & 62.1 {\scriptsize(+72/$-$47; 0.03)} & 32.4 {\scriptsize(+9/$-$3; 0.15)} \\
Free-form code (ledger only) & 61.1 {\scriptsize(+77/$-$46; 0.007)} & 40.5 {\scriptsize(+9/$-$6; 0.61)} \\
Free-form code, thinking (ledger + log) & 60.2 {\scriptsize(+82/$-$45; 0.001)} & 45.9 {\scriptsize(+9/$-$8; 1.00)} \\

\bottomrule
\end{tabular}
\caption{Free-form code control (Qwen3-32B, accuracy \%). Columns: the \CodeN{} rule-selected questions and those with
load above 100. In parentheses: questions that \sys{}
answers correctly and the code arm does not / the reverse, and the two-sided sign-test $p$.}
\label{tab:code}
\end{table*} 

For the primary arm, program generation cost \$\CodeGenCost{} and answering and
judging \$\CodeAnswerCost.

\section{MINTEval Protocol and Detailed Results}
\label{app:mint}

\begin{figure}[!tb]
\centering
\includegraphics[width=\columnwidth]{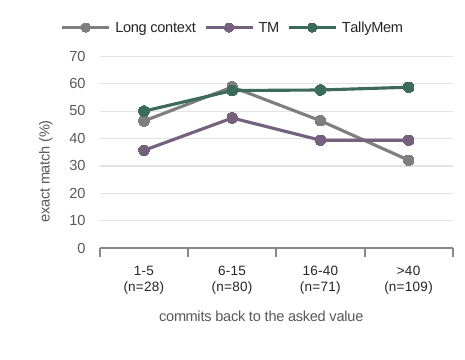}
\caption{MINTEval value-history questions by commits back to the requested value: 288 questions with an annotated
lookback of at least one commit (bin sizes shown).
\sys{} exceeds TM in every bin; the two farthest bins also show larger
advantages over long-context reading. The bins describe different questions, not a controlled intervention on
history length.}
\label{fig:mint_distance}
\end{figure}

The GitHub subset contains 200 repositories with 22--148 commits each. We use 25 of the 101 development repositories
for adaptation and freeze the system before evaluating 99 test repositories (\MintN{} questions). Each commit is a
step; the ledger is extracted from diffs, including file additions, deletions, renames and added assignment lines.
The schema description in the compiler prompt changes accordingly. This is a benchmark-specific adaptation of the
interface, not zero-adaptation transfer. Outputs follow MINTEval's official format and exact-match evaluation.
Long context was evaluated after the pre-registered test under the same 40,960-token model context limit,
using a 70:30 head--tail split when truncation was needed. Its input budget is the model limit minus 12,288 output
tokens, 300 safety tokens and fixed prompt text; TM instead uses its 26k-token retrieval budget with per-field
caps of 1,200 characters, or 4,000 for anchor steps. TM also truncates long diffs, so admitting all commit IDs does not imply complete commit coverage.

\begin{table*}[t]
\centering\small
\setlength{\tabcolsep}{4pt}
\begin{tabular}{lccccccc}
\toprule
System & All & Hist. & Order & Count & Simple & Multi-hop & vs.\ TM \\
\midrule
Long context & 39.2 & 45.8 & 19.0 & 52.5 & 47.1 & 33.6 & +123/$-$92 (0.041) \\
TM & 35.2 & 41.0 & 17.6 & 31.1 & 47.8 & 31.3 & -- \\
\sys$^\P$ & 49.0 & 56.6 & 43.1 & 47.5 & 56.6 & 31.3 & +124/$-$17 ($<$0.001) \\
\sys, no routing & 49.0 & 56.6 & 43.1 & 47.5 & 56.6 & 31.3 & +125/$-$18 ($<$0.001) \\
\sys, AMA load routing$^\S$ & 35.7 & -- & -- & -- & -- & -- & +5/$-$1 (0.22) \\

\bottomrule
\end{tabular}
\caption{MINTEval GitHub split, 99 held-out repositories (\MintN{} questions), exact match (\%), Qwen3-32B reader,
frozen system evaluated once. Long context: the whole history, head and tail truncated to the budget of the AMA-Bench
long-context arm, run after the pre-registered test as a reference row. Sign tests are paired with TM. $^\P$With the
coverage-aware router of Appendix~\ref{app:gate}. $^\S$The load-only router.}
\label{tab:mint_detail}
\end{table*}

\sys{} versus TM gives \MintWL{} ($\MintP$), and versus long context gives
\MintTallyVsLongWL{} ($\MintTallyVsLongP$). The gains are largest for file order
(\MintOrdTm\%$\to$\MintOrdTally\%) and value histories (\MintHistTm\%$\to$\MintHistTally\%).
Multi-hop performance is unchanged. On counting, long context scores \MintCountLong\% versus \MintCountTally\%
for \sys{} (not significant).

\paragraph{Other readers.}
After the pre-registered test we repeated the three arms with gpt-oss-120b (DeepInfra, the configuration of its
AMA-Bench column) and Qwen3-8B (local vLLM, thinking) as answer models. They reuse the Qwen3-32B compiler's evidence,
TM's contexts and the routing, and every prompt is byte-identical to the Qwen3-32B run, which we checked by hash before
any answer was generated. \sys{} exceeds TM with both readers (\MintOssTallyVsTmWL, $\MintOssTallyVsTmP$, and
\MintEightTallyVsTmWL, $\MintEightTallyVsTmP$; repository-clustered intervals \MintOssTallyVsTmCI{} and
\MintEightTallyVsTmCI{} points) and long context (\MintOssTallyVsLongWL, $\MintOssTallyVsLongP$, and
\MintEightTallyVsLongWL, $\MintEightTallyVsLongP$). TM again trails long context (\MintOssTmVsLongWL,
$\MintOssTmVsLongP$, and \MintEightTmVsLongWL, $\MintEightTmVsLongP$).

\section{Error Analysis Protocol}
\label{app:taxonomy}
The analysis covers the \ErrN{} errors of \sys{} on the \CodeN{} diagnostic questions with Qwen3-32B, using the
principal configuration with computed evidence always attached ($\gamma=1$). It uses frozen outputs only: no model
was called and no method was changed afterwards. The procedure was written before any error was inspected.

\paragraph{Stage 1.} Errors without computed evidence (\ErrNone) are separated automatically. Their answers are TM's:
the compiler declined in \ErrNoneNA{} cases, every sub-query failed the grounding checks in \ErrNoneDropped, and
compilation failed in the rest.

\paragraph{Stage 2.} Each remaining error was labelled from the question, the reference, the compiled plan, the
per-sub-query outcomes, the rendered evidence and the final answer, and the raw log and ledger were consulted whenever
the label depended on them. The first applicable label is assigned, in this order:
reference or judge (the answer agrees with the log in substance, and the reference contradicts the log or the judge
rejects a correct answer); answering (the rendered evidence states everything the reference requires, or the answer
contradicts what it states); partial coverage (the evidence is correct and the error lies only in parts answered from
TM's view); specification (the scope, predicate or operation does not express the referenced quantity, including a
requested quantity that was not compiled or was dropped); ledger (the plan expresses the quantity but ledger rows are
wrong or inconsistent); truncation (a correct result was cut by the rendering budget); unclear. A plan whose reading of
an ambiguous question is defensible is still a specification error and is flagged; \ErrSpecAlt{} of the \ErrSpec{}
specification errors carry this flag. Four clarifications were added while labelling and applied to all cases:
contradicting stated evidence counts as an answering error; evidence that must be joined or recomputed across
sub-queries does not state the quantity; a flawed reference yields the reference label only if the answer is correct
against the log; and an answer that adds a false claim to otherwise complete content is an answering error. The last
rule moved four cases out of the reference category.

\paragraph{Validity check.} The evidence of each question is shared by all readers, so a systematic error should
persist across readers while an answering error should not. As specified in advance, we compared the accuracy of the
other four readers on answering and specification errors: \ErrValA\% versus \ErrValS\% (two-sided
permutation test, $p=\ErrValP$). Of the \ErrSpec{} specification errors, \ErrSpecClear{} misread an unambiguous question and \ErrSpecAmb{} concern questions whose plan reading is defensible while the reference adopts another; the other readers answer \ErrValSClear\% and \ErrValSAmb\% of them correctly, so both groups are systematic.

\begin{table*}[t]
\centering\small
\begin{tabular}{lrrrrrrr}
\toprule
Questions & Errors & Specification & Ledger & Answering & Partial & Reference/judge & Unclear \\
\midrule
Load 1--15 & 86 & 24 & 13 & 8 & 9 & 30 & 2 \\
Load 16--100 & 43 & 17 & 8 & 10 & 2 & 6 & 0 \\
Load $>$100 & 19 & 10 & 0 & 3 & 0 & 6 & 0 \\
Count / frequency (F1) & 52 & 21 & 14 & 10 & 5 & 2 & 0 \\
First / last change (F2) & 29 & 10 & 1 & 2 & 3 & 12 & 1 \\
State tracing (F3) & 55 & 7 & 7 & 8 & 4 & 28 & 1 \\
Action enumeration (F4) & 18 & 15 & 0 & 1 & 2 & 0 & 0 \\

\bottomrule
\end{tabular}
\caption{Errors with computed evidence by aggregation load and question family (Qwen3-32B, \CodeN{} diagnostic
questions). A question can belong to several families. Executor and truncation errors: none.}
\label{tab:error_breakdown}
\end{table*}

\section{Error Analysis and Later Compiler Changes}
\label{app:errors}

\begin{figure*}[t]
\centering
\includegraphics[width=\textwidth]{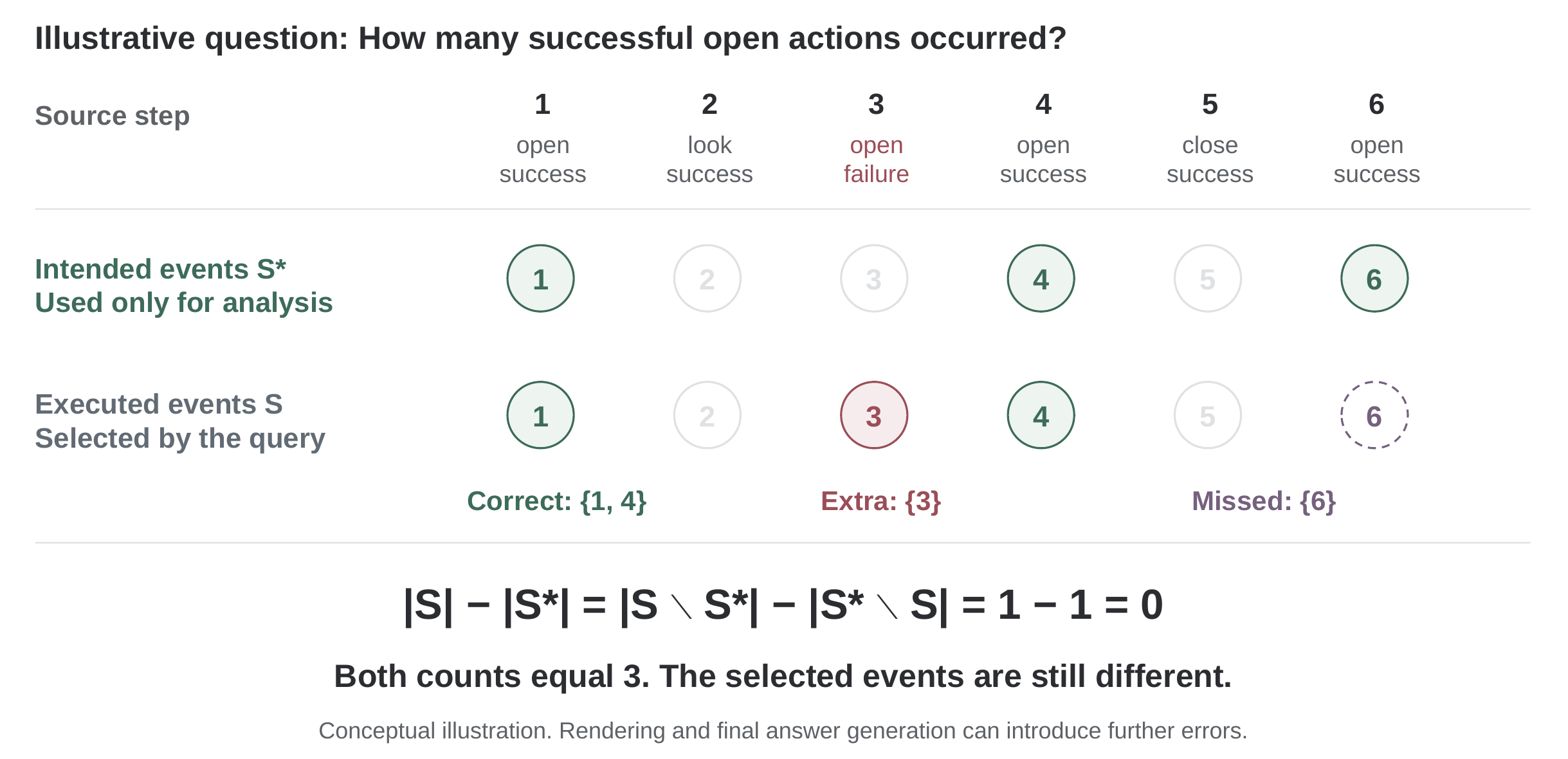}
\caption{A conceptual example of Equation~\ref{eq:selectionerror}. The executed set includes an extra event
and misses an intended one. Their effects cancel in the scalar count, although the selected sets differ.
$S^{\star}$ is an analysis reference, not an input to inference. Rendering and final answer generation can
introduce additional errors.}
\label{fig:selection_errors}
\end{figure*}

\paragraph{Selection errors.}
For a query that counts logged steps, let $S$ be the executed set and $S^{\star}$ the intended set of
Equation~\ref{eq:target}. Before rendering and answering, the computed count error is
\begin{equation}
 |S|-|S^{\star}|=|S\setminus S^{\star}|-|S^{\star}\setminus S|,
 \label{eq:selectionerror}
\end{equation}
the difference between extra and missed events. Both terms can be checked against the log because each result states
its filters and supporting step IDs, although $S^{\star}$ itself is unknown at inference time. Equal counts can still
conceal cancelling errors (Figure~\ref{fig:selection_errors}).

\paragraph{Where computation does not help.}
First/last-change and single-step state questions are flat (Table~\ref{tab:families}), because retrieval already finds
the relevant step. By domain, the gain comes mostly from embodied trajectories (Table~\ref{tab:domain} in Appendix~\ref{app:domain}). Multi-part text-to-SQL
statistics improve only partly (22 $\to$ 27 of 51), because the gold
answers' notion of an ``object'' often differs from the ledger's. Harmful evidence mostly comes from inconsistent target
labels in the annotated ledger (for example, a web element labelled differently across steps) and from content
questions about code, which the refined compiler now leaves to retrieval.

\paragraph{Execution fails when the quantity is misspecified.}
The following Qwen3-32B DeepInfra diagnostic examples show programs executing exactly what the plan asked, while the plan's notion of the quantity differs from the gold answer's.
\begin{itemize}
  \item For ``which containers has the agent interacted with and what state changes occurred'', the gold answer lists
  only open and close events. The plan grouped every step that visited or looked at an object.
  \item For ``which objects has the agent interacted with and how many times'', the gold answer counts only objects
  picked up or moved. The plan counted visits to furniture.
\end{itemize}
Sampling more plans does not fix a definition the compiler consistently gets wrong, which explains why
self-consistency fails. A general rule does fix part of it: questions about state changes use only steps with recorded
changes. The refined compiler raises the container template from 4 to 12 of 26 questions. Overall, the refined compiler answers 423 of
the 630 questions correctly, against 416 for the principal compiler and 390 for TM. The computation path thus converts an error that
cannot be inspected (a
miscounted list) into one that can be examined: a query that says what it counted.

\paragraph{A fixed counter is a special case.}
TM's keyword-triggered counter is itself a small program. On count questions where it fires, TM is nearly as accurate
as \sys{} (67\% vs.\ 68\%, 221 questions). A fixed aggregate works when the question happens to match it; the probe isolates this effect by asking the same
count in two wordings (Appendix~\ref{app:probe}).
Compiling the question extends this to groupings, outcome breakdowns, first/last queries and timelines.

\section{Results by Domain}
\label{app:domain}
\begin{table*}[t]
\centering\small
\setlength{\tabcolsep}{4pt}
\begin{tabular}{lcccccc}
\toprule
Domain & $n$ & Long ctx. & TM & \sys-lite & \sys & \sys{} vs.\ TM \\
\midrule
Embodied & 239 & 129 & 141 & 161 & 165 & +33/$-$9 ($<$0.001) \\
Game & 17 & 12 & 10 & 10 & 8 & +0/$-$2 (0.50) \\
Open-world & 59 & 30 & 45 & 45 & 43 & +0/$-$2 (0.50) \\
Software & 67 & 27 & 21 & 21 & 20 & +3/$-$4 (1.00) \\
Text-to-SQL & 153 & 97 & 99 & 96 & 107 & +16/$-$8 (0.15) \\
Web & 95 & 55 & 74 & 74 & 73 & +3/$-$4 (1.00) \\

\bottomrule
\end{tabular}
\caption{Correct answers by domain on the 630 rule-selected questions (Qwen3-32B). The gain is concentrated in the
embodied domain, whose trajectories are long and whose questions aggregate the most steps. Long context is better
than TM on software engineering and games (small samples), where TM's selection misses context.}
\label{tab:domain}
\end{table*}

\clearpage
\makeatletter
\@colht=\textheight
\@colroom=\textheight
\vsize=\textheight
\makeatother
\raggedbottom
\section{Worked Example: From Logged Steps to Computed Evidence}
\label{app:examples}
Episode 30 is an embodied trajectory with 150 steps. The question is ``Until step 149, what types of actions has the agent performed and how frequently?''.

\paragraph{Access and aggregation.}
TM's reading view shows 132 of the 150 steps, omitting the first 18. Even within the visible portion, its answer miscounts five of the eight action types; for example, it reports \texttt{examine} 18 times against a true 5. This example illustrates both evidence access and aggregation. The complete-context control is shown separately in Figure~\ref{fig:case}.

\newpage
\paragraph{Executable evidence.}
\sys{} compiles one sub-query with $I_1=[0,149]$, a predicate admitting every logged tool and $\omega_1=\texttt{group}_{\mathrm{tool}}$. Equation~\ref{eq:operations} computes the frequencies over all ledger rows in scope. The result is \texttt{go} 77, \texttt{look} 49, \texttt{open} 7, \texttt{close} 5, \texttt{examine} 5, \texttt{help} 4, \texttt{take} 2 and \texttt{move} 1, with supporting step IDs (Figure~\ref{fig:example}). The reader reproduces these counts.

\begin{figure}[H]
\centering
\fbox{\parbox{0.96\columnwidth}{\ttfamily\scriptsize
[Q1] group\_count by tool; filter: tool any of ['go','look','open','close',\\
\hspace*{2em}'examine','help','take','move']; matched steps: 150\\
\hspace*{1em}go: 77 (steps 1-7, 10-11, 14, ...) - outcomes: success 77\\
\hspace*{1em}look: 49 (steps 0, 15, 26, ...) - outcomes: success 32, no\_effect 17\\
\hspace*{1em}open: 7 (steps 8, 12, 19, 23, 61, 85, 109) - outcomes: success 7\\
\hspace*{1em}...}}
\caption{Computed evidence $D_q$: tool frequencies, supporting step IDs and outcome breakdowns (excerpt).}
\label{fig:example}
\end{figure}

\begin{figure*}[t]
\centering
\includegraphics[width=\textwidth]{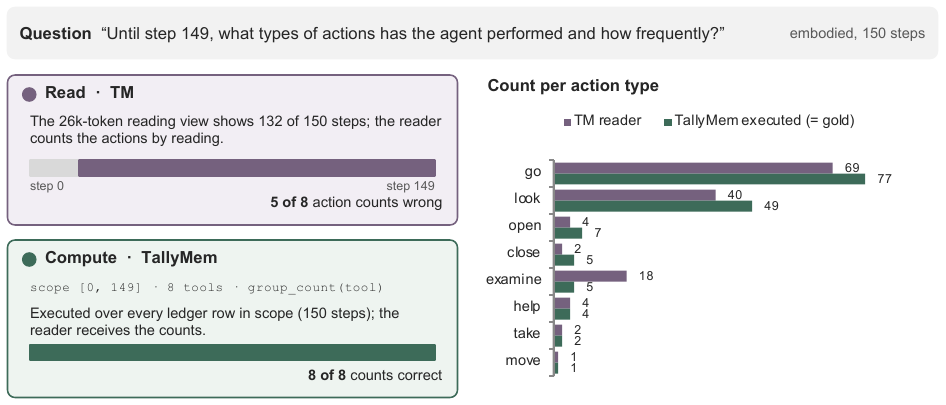}
\caption{Reading and execution on the worked example. Left: TM's reading view and the full-ledger computation path. Right: saved TM counts and the executable result against the reference.}
\label{fig:tm_count_case}
\end{figure*}

\begin{figure*}[t]
\centering
\includegraphics[width=\textwidth,trim=0 18 0 0,clip]{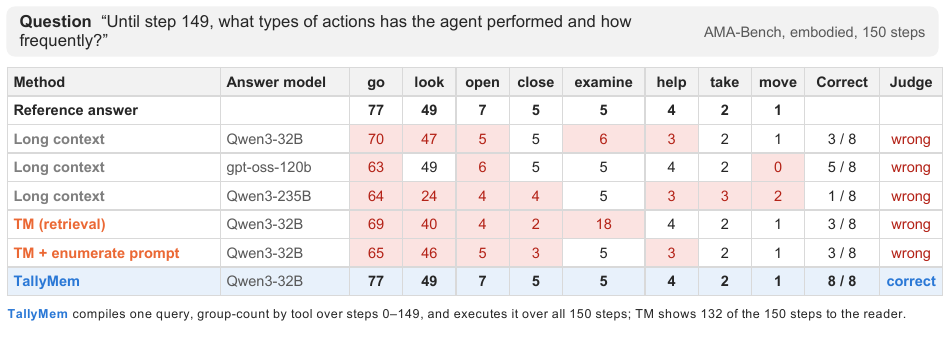}
\caption{Saved answers across reader configurations on the same question. Red entries differ from the reference; all six answer rows come from diagnostic runs.}
\label{fig:case_answers}
\end{figure*}

\paragraph{Saved answers.}
Figure~\ref{fig:tm_count_case} contrasts TM's reading view with the executable evidence; Figure~\ref{fig:case_answers} compares saved answers across reader configurations. Unlike the illustrative memory cards in Figure~\ref{fig:case}, these outputs come from the diagnostic runs.

\clearpage

\clearpage

\end{document}